\documentclass[runningheads]{llncs}
\usepackage{graphicx}
\usepackage{amsmath,amssymb} % define this before the line numbering.
\usepackage{color}
\usepackage[width=122mm,left=12mm,paperwidth=146mm,height=193mm,top=12mm,paperheight=217mm]{geometry}

\usepackage{epsfig}
\usepackage{afterpage}
\usepackage{multirow}
\usepackage{subfigure}
\usepackage{comment}
\usepackage{microtype}
\usepackage{float}
\usepackage[toc,page]{appendix}
\usepackage{soul}
\usepackage[dvipsnames]{xcolor}

\usepackage{capt-of}

\usepackage{enumitem}
\setitemize[0]{leftmargin=10pt}

\newenvironment{tight_itemize}{
\begin{itemize}[leftmargin=10pt]
  \setlength{\topsep}{0pt}
  \setlength{\itemsep}{0pt}
  \setlength{\parskip}{0pt}
  \setlength{\parsep}{0pt}
}{\end{itemize}}

\newcommand{\KS}[1]{\textcolor{red}{\textbf{Kalyan: #1}}}
\newcommand{\ZL}[1]{\textcolor{blue}{\textbf{Zhengqin: #1}}}

\begin{document}

\pagestyle{headings}
\mainmatter
\def\ECCV18SubNumber{2455}  % Insert your submission number here

%%%%%%%%% TITLE
%\title{Single Image BRDF Reconstruction Using Convolutional Network}
%\title{Single-Image SVBRDF Estimation Using a Consumer Mobile Phone}
\title{Materials for Masses: SVBRDF Acquisition with a Single Mobile Phone Image}

\titlerunning{SVBRDF Acquisition with a Single Mobile Phone Image}

\authorrunning{Li, Sunkavalli and Chandraker}

\author{Zhengqin Li{$^*$} {\hspace{1cm}} Kalyan Sunkavalli{$^\dagger$} {\hspace{1cm}} Manmohan Chandraker{$^*$}}
\institute{{$^*$}University of California, San Diego {\hspace{1cm}} {$^\dagger$}Adobe Research, San Jose}

\maketitle

\begin{abstract}
\vspace{-0.3cm}
We propose a material acquisition approach to recover the spatially-varying BRDF and normal map of a near-planar surface from a single image captured by a handheld mobile phone camera. Our method images the surface under arbitrary environment lighting with the flash turned on, thereby avoiding shadows while simultaneously capturing high-frequency specular highlights. We train a CNN to regress an SVBRDF and surface normals from this image. Our network is trained using a large-scale SVBRDF dataset and designed to incorporate physical insights for material estimation, including an in-network rendering layer to model appearance and a material classifier to provide additional supervision during training. We refine the results from the network using a dense CRF module whose terms are designed specifically for our task.
%We demonstrate that our CNN-based SVBRDF inference
The framework is trained end-to-end and produces high quality results for a variety of materials. We provide extensive ablation studies to evaluate our network on both synthetic and real data, while demonstrating significant improvements in comparisons with prior works.

\end{abstract}

\vspace{-0.4cm}
\section{Introduction}
\label{sec:intro}
\vspace{-0.2cm}

The wide variety of images around us are the outcome of interactions between lighting, shapes and materials. In recent years, the advent of convolutional neural networks (CNNs) has led to significant advances in recovering shape using just a single image \cite{eigenNet,3D2N2}. In contrast, material estimation has not seen as much progress, which might be attributed to multiple causes. First, material properties can be more complex. Even discounting more complex global illumination effects, materials are represented by a spatially-varying bidirectional reflectance distribution function (SVBRDF), which is an unknown high-dimensional function that depends on exitant and incident lighting directions \cite{nicodemus}. Second, while large-scale synthetic and real datasets have been collected for shape estimation \cite{shapenet,nyudataset}, there is a lack of similar data for material estimation. Third, pixel observations in a single image contain entangled information from factors such as shape and lighting, besides material, which makes estimation ill-posed.

In this work, we present a practical material capture method that can recover an SVBRDF from a \emph{single} image of a near-planar surface, acquired using the camera of an off-the-shelf consumer mobile phone, under unconstrained environment illumination. This is in contrast to conventional BRDF capture setups that usually require significant equipment and expense \cite{acquiringBRDF1,acquiringBRDF3}. We address this challenge by proposing a novel CNN architecture that is specifically designed to account for the physical form of BRDFs and the interaction of light with materials, which leads to a better learning objective. We also propose to use a novel dataset of SVBRDFs that has been designed for perceptual accuracy of materials. This is in contrast to prior datasets that are limited to homogeneous materials, or juxtapose material properties with other concepts such as object categories. 

\begin{figure}[!!t]
\begin{center}
\begin{minipage}[t]{0.55\linewidth}
\raisebox{-2.8cm}
{
\begin{tabular}{ccc}
\includegraphics[width=0.75in]{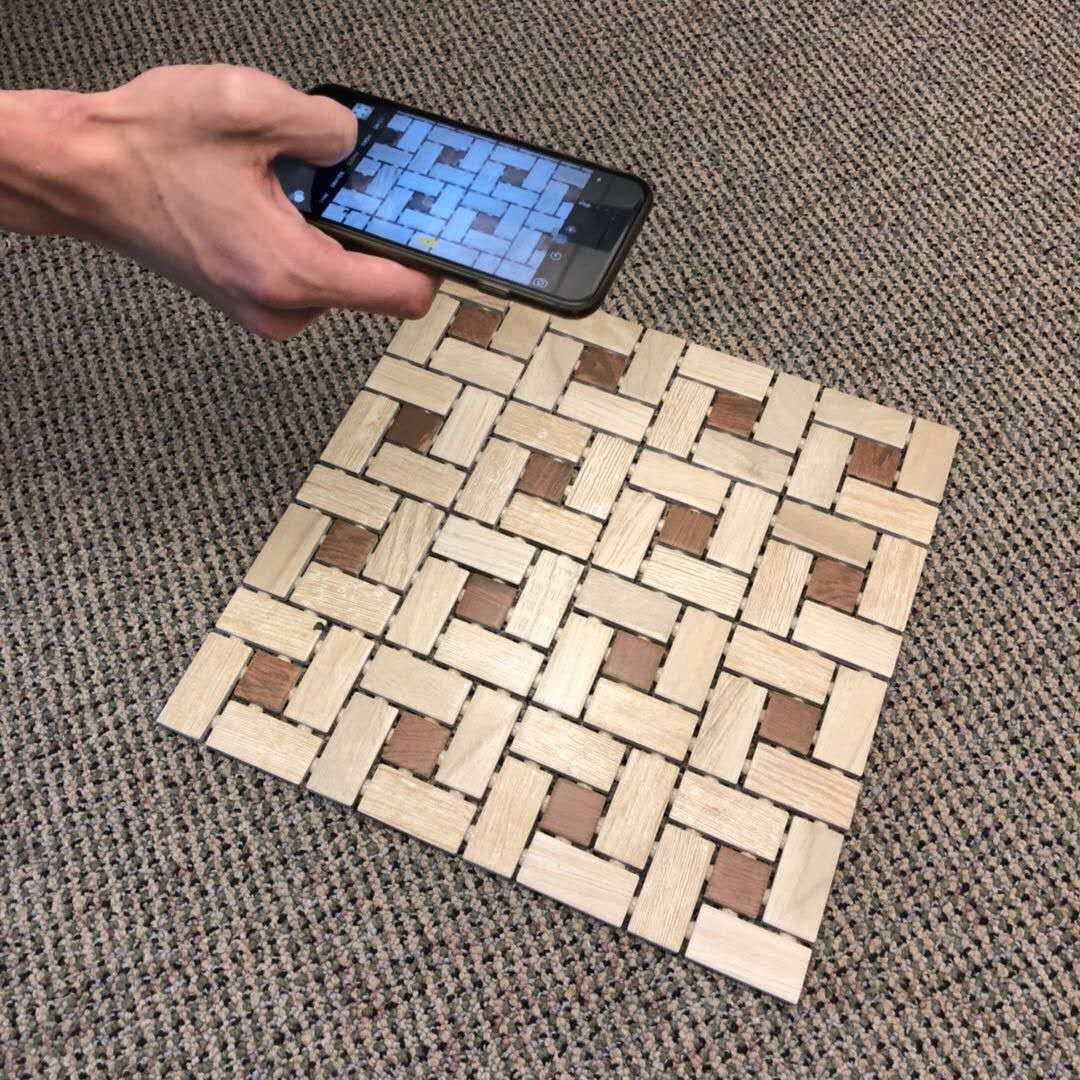} & 
\includegraphics[width=0.75in]{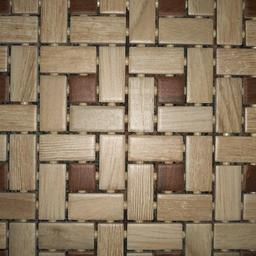} & 
\includegraphics[width=0.75in]{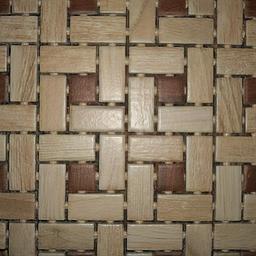}\\[-0.1cm]
{\scriptsize Setup} & {\scriptsize Input} & {\scriptsize Rendering}\\
\includegraphics[width=0.75in]{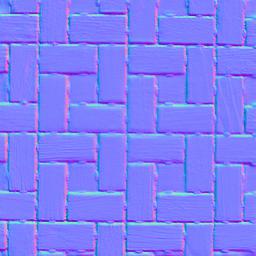} & 
\includegraphics[width=0.75in]{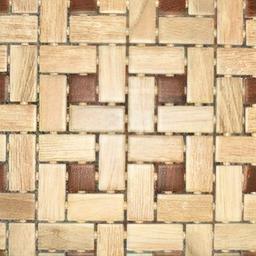} &
\includegraphics[width=0.75in]{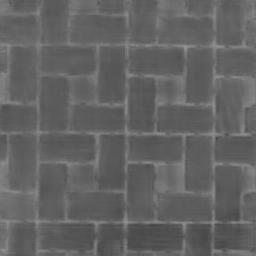} \\[-0.1cm]
{\scriptsize Normal} & {\scriptsize Albedo} & {\scriptsize Roughness} 
\end{tabular}
}
\end{minipage}\hfill
\begin{minipage}[t]{0.45\linewidth}
\caption{\small
We propose a deep learning-based light-weight SVBRDF acquisition system. From a single image of a near planar surface captured with a flash-enabled mobile phone camera under arbitrary lighting, our network recovers surface normals and spatially-varying BRDF parameters -- diffuse albedo and specular roughness. Rendering the estimated parameters produces an image almost identical to the input image.
\label{teaser}
}
\end{minipage}
\end{center}
\vspace{-1.2cm}
\end{figure}

%% \begin{figure}
%% \centering
%% \includegraphics[width=2.9in]{teaser.jpg}
%% \caption{We propose a deep learning-based light-weight BRDF acquisition system. From a single image (a) of a near planar surface  captured with a mobile phone camera under flash illumination, our network recovers spatially-varying BRDF parameters --- diffuse albedo (b), and specular roughness (e) --- and surface normals (c). Rendering the estimated parameters produces an image (d) that is almost identical to the input image.}
%% \label{teaser}
%% \vspace{-0.5cm}
%% \end{figure}

%In Section \ref{sec:method} 
We introduce a novel CNN architecture that encodes the input image into a latent representation, which is decoded into components corresponding to surface normals, diffuse texture, and specular roughness. We propose a differentiable rendering layer that recombines the estimated components with a novel lighting direction. This gives us additional supervision from images of the material rendered under arbitrary lighting directions during training; only a single image is used at test time. We also observe that coarse classification of BRDFs into material meta-categories is an easier task, so we additionally include a material classifier to constrain the latent representation. The inferred BRDF parameters from the CNN are quite accurate, but we achieve further improvement using densely-connected conditional random fields (DCRFs) with novel unary and smoothness terms that reflect the properties of the underlying microfacet BRDF model. We train the entire framework in an end-to-end manner.

Our approach --- using our novel architecture and SVBRDF dataset --- can outperform the state-of-art. We demonstrate that we can further improve these results by leveraging a form of acquisition control that is present on virtually every mobile phone --- the camera flash. We turn on the flash of the mobile phone camera during acquisition; our images are thus captured under a combination of unknown environment illumination and the flash. The flash illumination helps further improve our reconstructions. First, it minimizes shadows caused by occlusions. Second, it allows better observation of high-frequency specular highlights, which allows better characterization of material type and more accurate estimation. Third, it provides a relatively simple setup for acquisition that eases the burden on estimation and allows the use of better post-processing techniques. 

In contrast to recent works such as \cite{twoshotBRDF} and \cite{styleBRDF} that can reconstruct BRDFs with stochastic textures, we can handle a much larger class of materials. Also, our results, both with and without flash, are a significant improvement over the recent method of Li et al. \cite{CNN-BRDF} even though our trained model is more compact. ~%try to reconstruct a BRDF under an unconstrained environment using data-driven based methods in \cite{CNN-BRDF}. However, our experiments in Section \ref{sec:experiments} show that reconstructing BRDFs from purely environment lighting is outperformed by a setup where point-light input is additionally available.
Our experiments demonstrate advantages over several baselines and prior works in quantitative comparisons, while also achieving superior qualitative results. In particular, the generalization ability of our network trained on the synthetic BRDF dataset is demonstrated by strong performance on real images, acquired in the wild, in both indoor and outdoor environments, using multiple different phone cameras. Given the estimated BRDF parameters, we also demonstrate applications such as material editing and relighting of novel shapes.
%\st{Note that our single-image setting allows material estimation and editing at an arbitrary point in time after the image is recorded, even for images not explicitly acquired for the purpose.} \KS{we do need flash illumination so not quite right?} \ZL{You are definitely correct.@KS I think we can remove this sentence.}

To summarize, we propose the following contributions:
\vspace{-0.2cm}
\begin{tight_itemize}
%\item A new SVBRDF acquisition framework that balances acquisition effort and reconstruction quality. 
%\item Demonstration of accurate SVBRDF estimation from a single image captured with a consumer mobile phone. 
\item A novel lightweight SVBRDF acquisition method that produces state-of-the-art reconstruction quality. 
\item A CNN architecture that exploits domain knowledge for joint SVBRDF reconstruction and material classification.
\item Novel DCRF-based post-processing that accounts for the microfacet BRDF model to refine network outputs.
\item An SVBRDF dataset that is large-scale and specifically attuned to estimation of spatially-varying materials.
\end{tight_itemize}

\section{Related Work}
\label{sec:related}

%In this section, we briefly review previous work on deep learning based intrinsic image decomposition and lightweight BRDF acquisition method. 

\noindent\textbf{BRDF Acquisition:} The Bidirectional Reflection Distribution function (BRDF) is a 4-D function that characterizes how a surface reflects lighting from an incident direction toward an outgoing direction~\cite{nicodemus}. Alternatively, BRDFs are represented using low-dimensional parametric models~\cite{blinn1976texture,cook1982reflectance,ward1992measuring,oren1995generalization}. In this work, we use a physically-based microfacet model~\cite{burley2012disney} that our SVBRDF dataset uses.

Traditional methods for BRDF acquisition rely on densely sampling this 4-D space using expensive, calibrated acquisition systems~\cite{acquiringBRDF1,acquiringBRDF3,matusik2003data}. Recent work has demonstrated that assuming BRDFs lie in a low-dimensional subspace allows for them to be reconstructed from a small set of measurements~\cite{opDirecBRDF1,opDirecBRDF2}. However, these measurements still to be taken under controlled settings. We assume a single image captured under largely uncontrolled settings.

Photometric stereo-based methods recover shape and/or BRDFs from images. Some of these methods recover a homogeneous BRDF given one or both of the shape and illumination~\cite{romeiro2008,romeiro2010,oxholm2016shape}. Chandraker et al. \cite{chandraker1,chandraker2,chadraker3} utilize motion cues to jointly recover the shape and BRDF of objects from images under known directional illumination. Hui et al.~\cite{hui2015dic} recover SVBRDFs and shape from multiple images under known illuminations. All of these methods require some form of calibrated acquisition; in contrast, we wish to capture SVBRDFs and normal maps ``in-the-wild''.

Recent work has shown promising results for ``in-the-wild'' BRDF acquisition. Hui et al.~\cite{mobileBRDF} demonstrate that the collocated camera-light setup on mobile devices is sufficient to reconstuct SVBRDFs and normals. They need capture 30+ images and calibrate them to reconstruct SVBRDFs; we aim to do this from a single image. Aittala et al.~\cite{twoshotBRDF} propose using a flash/no-flash image pair to reconstruct \emph{stochastic} SVBRDFs and normals using a slow optimization-based scheme. Our method can handle a larger class of materials and is orders of magnitude faster.

\noindent\textbf{Deep learning-based Material Estimation:} 
Inspired by the success of deep learning for a variety of vision and graphics tasks, recent work has looked at CNN-based material recognition and estimation. Bell et al.~\cite{minc} train a material parsing network using crowd-sourced labeled data. However, their material recongition is driven more by object context, rather than appearance. Liu et al.~\cite{liu2017material} demonstrate image-based material editing using a network trained to recover homogenous BRDFs. Methods have been proposed to decompose images into their intrinsic image components which are an intermediate representation for material and shape~\cite{directIntrinsic,sceneIntrinsic,shapenetIntrinsics}. Rematas et al.~\cite{konsRefl} train a CNN to reconstruct the reflectance map -- a convolution of the BRDF with the illumination -- from a single image of a shape from a known class. In subsequent work, they disentangle the reflectance map into the BRDF and illumination~\cite{konsIlluRefl}. Neither of these methods handle SVBRDFs, nor do they recover fine surface normal details. Kim et al.~\cite{lwRefl} reconstruct a homegeneous BRDF by training a network to aggregate multi-view observations of an object of known shape .

Similar to us, Aittala et al.~\cite{styleBRDF} and Li et al.~\cite{CNN-BRDF} reconstruct SVBRDFs and surface normals from a single image of a near-planar surface. Aittala et al.~use a neural style transfer-based optimization approach to iteratively estimate BRDF parameters, however, they can only handle stationary textures and there is no correspondence between the input image and the reconstructed BRDF \cite{styleBRDF}. Li et al.~use supervised learning to train a CNN to predict SVBRDF and normals from a single image captured under environment illumination \cite{CNN-BRDF}. Their training set is small, which necessitates a self-augmentation method to generate training samples from unlabeled real data. Further, they train a different set of networks for each parameter (diffuse texture, normals, specular albedo and roughness) and each material type (wood, metal, plastic). We demonstrate that by using our novel CNN architecture, supervised training on a high-quality dataset and acquisition under flash illumination, we are able to (a) reconstruct all these parameters with a single network, (b) learn a latent representation that also enables material recognition and editing, (c) produce results that are significantly better qualitatively and quantitatively.

\section{Acquisition Setup and SVBRDF Dataset}
\label{sec:dataset}

In this section, we describe the setup for single image SVBRDF acquisition and the dataset we use for learning.

\vspace{-0.4cm}
\paragraph{Setup}
Our goal is to reconstruct the spatially-varying BRDF of a near planar surface from a single image captured by a mobile phone with the flash turned on for illumination. We assume that the $z$-axis of the camera is approximately perpendicular to the planar surface (we explicitly evaluate against this assumption in our experiments). For most mobile devices, the position of the flash light is usually very close to the position of the camera, which provides us a univariate sampling of a isotropic BRDF~\cite{mobileBRDF}. We argue that by imaging with a collocated camera and point light, we can have additional constraints that yield better BRDF reconstructions compared to acquisition under just environment illumination.

Our surface appearance is represented by a microfacet parametric BRDF model~\cite{burley2012disney}. Let $\mathbf{d}_{i}$, $\mathbf{n}_{i}$, $r_{i}$ be the diffuse color, normal and roughness, respectively, at pixel $i$. Our BRDF model is defined as: 
\begin{equation}
\rho(\mathbf{d}_{i}, \mathbf{n}_{i}, r_{i}) = \mathbf{d}_{i} + 
\frac{D(\mathbf{h}_{i}, r_{i})F(\mathbf{v}_{i}, \mathbf{h}_{i})G(\mathbf{l}_{i}, \mathbf{v}_{i}, \mathbf{h}_{i}, r_{i})}{4(\mathbf{n}_{i} \cdot \mathbf{l}_{i})(\mathbf{n}_{i}\cdot \mathbf{v}_{i})}
\label{eqn:brdf-model}
\end{equation}
where $\mathbf{v}_{i}$ and $\mathbf{l}_{i}$ are the view and light directions and $\mathbf{h}_{i}$ is the half angle vector. Given an observed image $I(\mathbf{d}_{i}, \mathbf{n}_{i}, r_{i}, \mathbf{L})$, captured under unknown illumination $\mathbf{L}$, we wish to recover the parameters $\mathbf{d}_{i}$, $\mathbf{n}_{i}$ and $r_{i}$ for each pixel $i$ in the image. Please refer to the supplementary material for more details on the BRDF model.

\begin{figure}[!!t]
  \begin{minipage}[b]{0.46\linewidth}
    \centering
    \begin{tabular}{cccc}
      \includegraphics[width=0.5in]{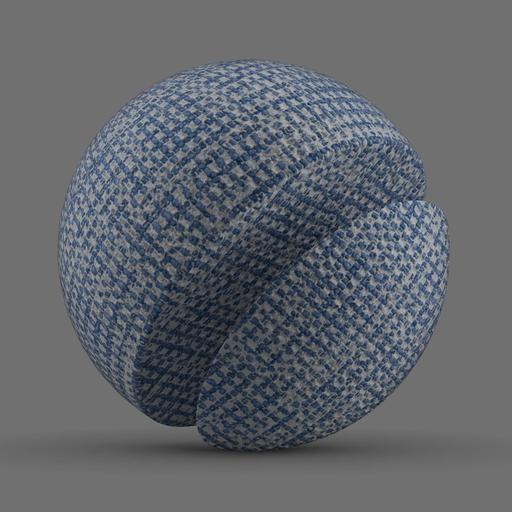} & 
      \includegraphics[width=0.5in]{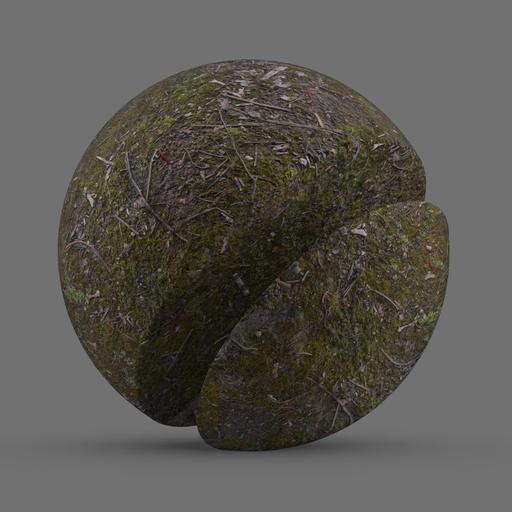} & 
      \includegraphics[width=0.5in]{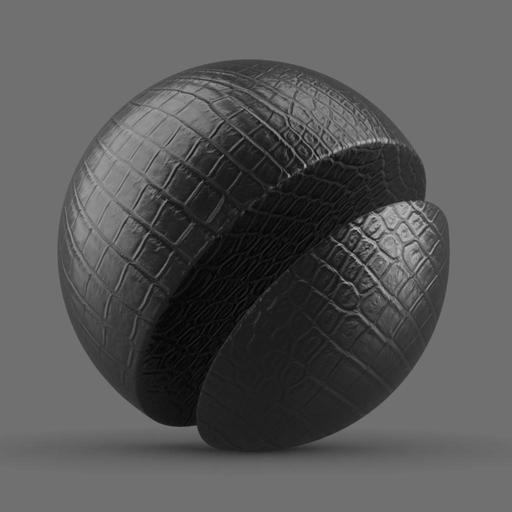} & 
      \includegraphics[width=0.5in]{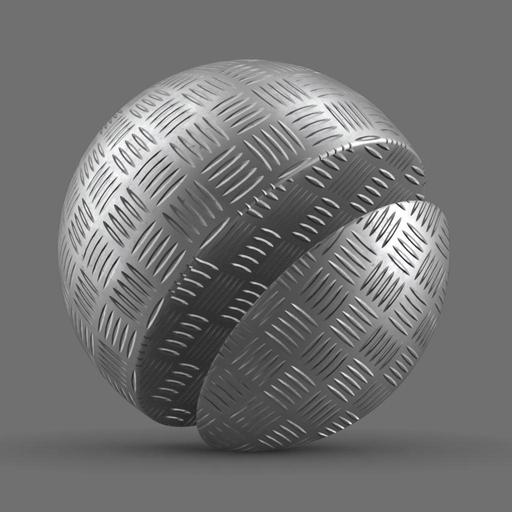} \\[-0.2cm]
      {\scriptsize fabric} & {\scriptsize ground} & {\scriptsize leather} & {\scriptsize metal} \\
      \includegraphics[width=0.5in]{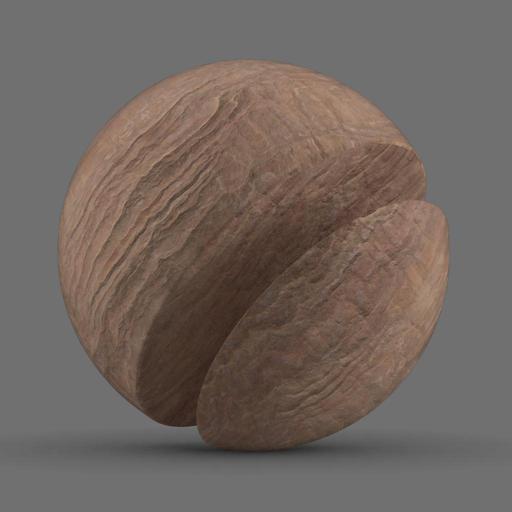} & 
      \includegraphics[width=0.5in]{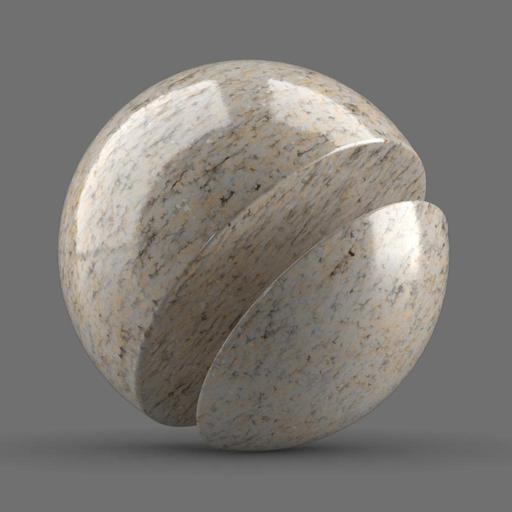} & 
      \includegraphics[width=0.5in]{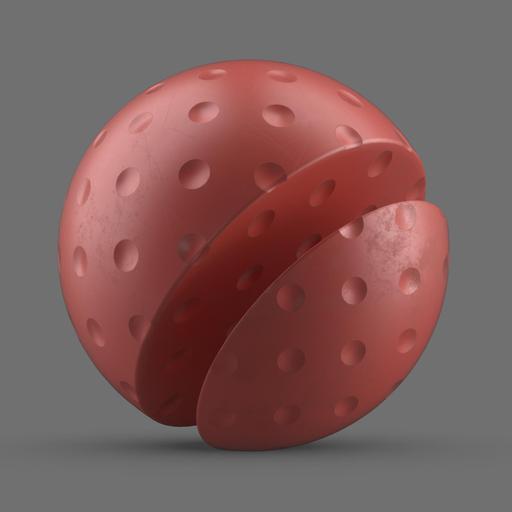} & 
      \includegraphics[width=0.5in]{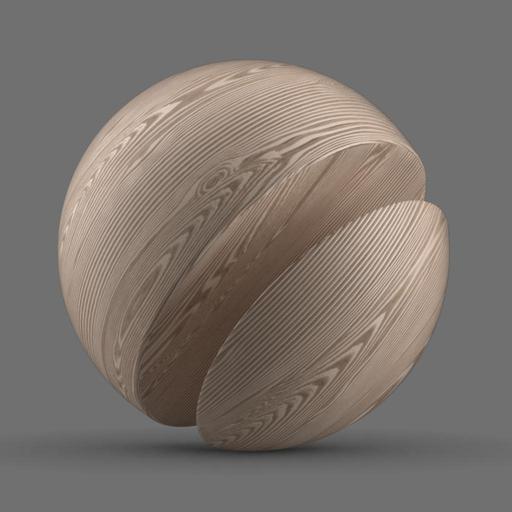} \\[-0.2cm]
      {\scriptsize stone-diff} & {\scriptsize stone-spec} & {\scriptsize polymer} & {\scriptsize wood} \\
    \end{tabular}
    \vspace{-0.3cm}
    \caption{
      %Representative examples from the 8 material types in the Adobe Stock dataset rendered on a sphere.
      Examples of our material types.
    }
    \label{fig:SVBRDF}
  \end{minipage}\hfill
  \begin{minipage}[b]{0.50\linewidth}
    \centering
    \begin{tabular}{|lcc|lcc|}
      \hline
      Materials & Train & Test 	& Materials & Train & Test \\
      \hline
      fabric & 165 & 29 & polymer 	& 33 		& 6 \\
      ground 		& 23 		& 4   	& stone-diff 	& 177 	& 30 \\
      leather 	& 10 		& 2  		& stone-spec 	& 38 		& 6 \\
      metal 		& 82 		& 13    & wood   					& 60 		& 10 \\
      \hline
    \end{tabular}    
    \captionof{table}{Distribution of materials in our training and test sets.}
    \label{matDist}
  \end{minipage}
\end{figure}

\vspace{-0.2cm}
\paragraph{Dataset}
We train our network on the Adobe Stock 3D Material dataset\footnote{\url{https://stock.adobe.com/3d-assets}}, which contains 688 materials with high resolution ($4096 \times 4096$) spatially-varying BRDFs. Part of the dataset is created by artists while others are captured using a scanner.
%by densely sampling from different views and light directions. 
We use 588 materials for training and 100 materials for testing. For data augmentation, we randomly crop 12, 8, 4, 2, 1 image patches of size 512, 1024, 2048, 3072, 4096. We resize the image patches to a size of $256 \times 256$ for processing by our network. We flip patches along $x$ and $y$ axes and rotate them in increments of 45 degrees. Thus, for each material type, we have 270 image patches.\footnote{The total number of image patches for each material can be computed as $(12+8+4+2+1)\times(1 + 2 + 7) = 270$.} We randomly scale the diffuse color, normal and roughness for each image patch to prevent the network from overfitting and memorizing the materials. We manually segment the dataset into $8$ materials types. The distribution is shown in Table \ref{matDist}, with an example visualization of each material type in Figure \ref{fig:SVBRDF}. More details on rendering the dataset are in supplementary material. 
%\KS{something about how we render these materials?} \ZL{Yes}

\section{Network Design for SVBRDF Estimation}
\label{sec:method}

\begin{figure}[!!t]
  \centering
  \includegraphics[width=0.90\textwidth]{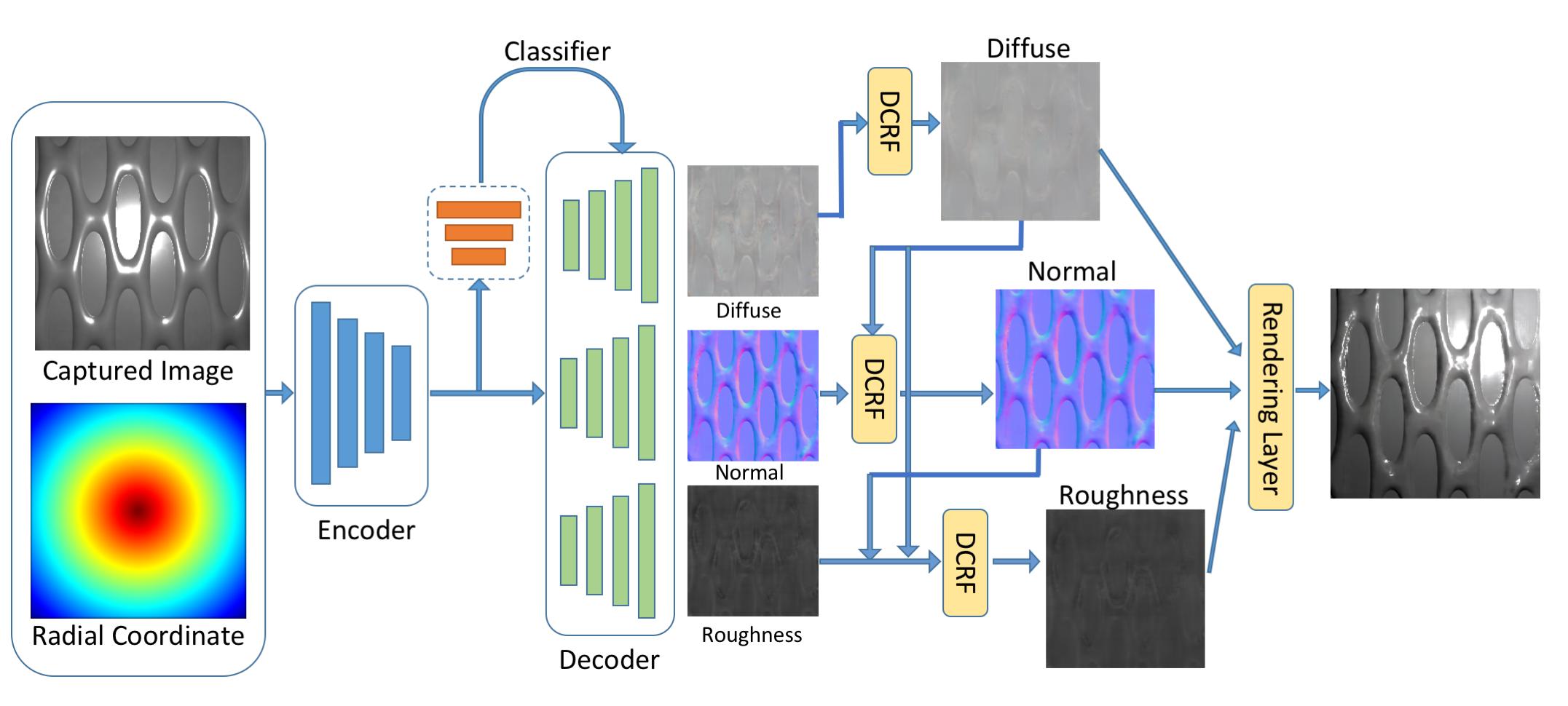}
  \vspace{-0.5cm}
  \caption{
    \small Our network for SVBRDF estimation consists of an encoder, three decoder blocks with skip links to retrieve SVBRDF components, a rendering layer and a material classifier, followed by a DCRF for refinement (not visualized). See Section \ref{sec:method} for how our architectural choices are influenced by the problem structure of SVBRDF estimation and supplementary material for the hyperparameter details.
  }
  \label{fig:network}
  \vspace{-0.2cm}
\end{figure}

In this section, we describe the components of our CNN designed for single-image SVBRDF estimation. The overall architecture is illustrated in Figure \ref{fig:network}.

\vspace{-0.3cm}
\subsection{Considerations for Network Architecture}
\vspace{-0.1cm}

Single-image SVBRDF estimation is an ill-posed problem. Thus, we adopt a data-driven approach with a custom-designed CNN that reflects physical intuitions.

Our basic network architecture consists of a single encoder and three decoders which reconstruct the three spatially-varying BRDF parameters: diffuse color $\mathbf{d}_{i}$, normals $\mathbf{n}_{i}$ and roughness $r_{i}$. The intuition behind using a single encoder is that different BRDF parameters are correlated, thus, representations learned for one should be useful to infer the others, which allows significant reduction in the size of the network. The input to the network is an RGB image, augmented with the pixel coordinates as a fourth channel. We add the pixel coordinates since the distribution of light intensities is closely related to the location of pixels, for instance, the center of the image will usually be much brighter. Since CNNs are spatially invariant, we need the extra signal to let the network learn to behave differently for pixels at different locations. Skip links are added to connect the encoder and decoders to preserve details of BRDF parameters.

Another important consideration is that in order to model global effects over whole images like light intensity fall-off or large areas of specular highlights, it is necessary for the network to have a large receptive field. To this end, our encoder network has seven convolutional layers of stride 2, so that the receptive field of every output pixel covers the entire image.

\vspace{-0.3cm}
\subsection{Loss Functions for SVBRDF Estimation}
\vspace{-0.1cm}

For each BRDF parameter, we have an L2 loss for direct supervision. We now describe other losses for learning a good representation for SVBRDF estimation.

\vspace{-0.2cm}
\paragraph{Rendering layer}
Since our eventual goal is to model the surface appearance, it is important to balance the contributions of different BRDF parameters. Therefore, we introduce a differentiable rendering layer 
%\KS{equation from this rendering layer in supplementary?}\ZL{Yes, that's the BRDF model we use to render images.} 
that renders our BRDF model (Eqn.~\ref{eqn:brdf-model}) under the known input lighting. We add a reconstruction loss based on the difference between these renderings with the predicted parameters and renderings with ground-truth BRDF parameters. The gradient can be backpropagated through the rendering layer to train the network. In addition to rendering the image under the input lighting, we also render images under by \emph{novel} lights. For each batch, we create novel lights by randomly sampling the the point light source on the upper hemisphere. 
%\KS{are these directional lights or point lights like the input?}\ZL{These are point light source}. 
This ensures that the network does not overfit to collocated illumination and is able to reproduce appearance under other light conditions. The final loss function for the encoder-decoder part of our network is:
\begin{equation}
\mathcal{L} = \lambda_{d}\mathcal{L}_{d} + \lambda_{n}\mathcal{L}_{n} + \lambda_{r}\mathcal{L}_{r} + \lambda_{rec}\mathcal{L}_{rec} ,
\label{energyBasic}
\end{equation}
where $\mathcal{L}_{d}$, $\mathcal{L}_{n}$, $\mathcal{L}_{r}$ and $\mathcal{L}_{rec}$ are the L2 losses for diffuse, normal, roughness and rendered image predictions, respectively. Here, $\lambda$'s are positive coefficients to balance the contributions of various terms, which are set to $1$ in our experiments.

Since we train on near planar surfaces, the majority of the normal directions are flat. Table \ref{normalDist} shows the normal distributions in our dataset. To prevent the network from over-smoothing the normals, we group the normal directions into different bins and for each bin we assign a different weight when computing the L2 error. This balance various normal directions in the loss function. 

\begin{table}[!!t]
  \scriptsize
\begin{minipage}[c]{0.45\textwidth}
\begin{tabular}{c}
\includegraphics[width=2.0in]{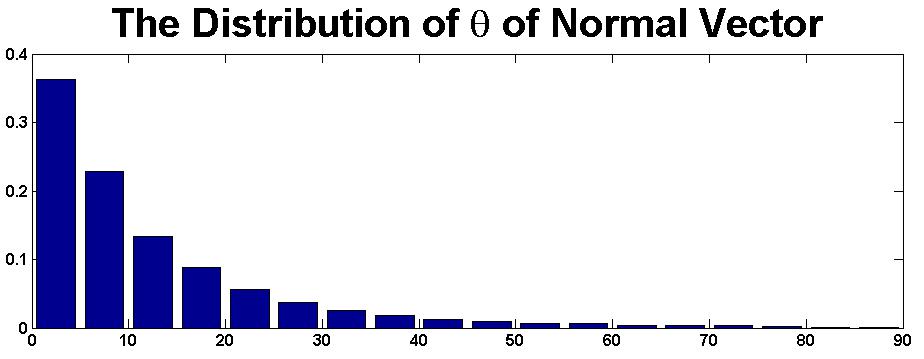} \\
\end{tabular}
\begin{tabular}{cccc}
\hline
Angle & $0^{\circ}\!-\!10^{\circ}$ & $10^{\circ}\!-\!25^{\circ}$  & $25^{\circ}\!-\!90^{\circ}$ \\
\hline
Prob($P_{i}$) & 0.592 & 0.278 & 0.130 \\
Weight($W_{i}$) & 0.869 & 1.060 & 1.469 \\
\hline
\end{tabular}
\end{minipage}\hfill
\begin{minipage}[c]{0.50\textwidth}
\caption{The $\theta$ distribution of the normal vector in the dataset, where $\theta$ is the angle between normal vector and $z$ axis. To avoid the network from over-smoothing the normal map, we group normal vectors into three bins according to $\theta$. With probability $P_{i}$ for bin $i$, its weight is $W_{i} = 0.7 + 1/10P_{i}$.} 
\label{normalDist}
\end{minipage}
\vspace{-0.3cm}
\end{table}

\vspace{-0.2cm}
\paragraph{Material Classification}
The distribution of BRDF parameters is closely related to the surface material type. However, training separate networks for different material types similar to~\cite{CNN-BRDF} is expensive. Also the size of the network grows linearly with the number of material types, which limits utility. Instead, we propose a split-merge network with very little computational overhead.

Given the highest level of features extracted by the encoder, we send the feature to a classifier to predict its material type. Then we evaluate the BRDF parameters for each material type and use the classification results as weights (the output of softmax layer). This averages the prediction from different material types to obtain the final BRDF reconstruction results. Suppose we have $N$ channels for BRDF parameters and $K$ material types. To output the BRDF reconstruction for each type of material, we only modify the last convolutional layer of the decoder so that the output channel will be $K\times N$ instead of $N$. In practice, we set $K$ to be $8$, as shown in Table \ref{matDist}.

The classifier is trained together with the encoder and decoder from scratch, with the weights of each label set to be inversely proportional to the number of examples in Table \ref{matDist} to balance different material types in the loss function. The overall loss function of our network with the classifier is  
\begin{equation}
\mathcal{L} = \lambda_{d}\mathcal{L}_{d} + \lambda_{n}\mathcal{L}_{n} + \lambda_{r}\mathcal{L}_{r} + \lambda_{rec}\mathcal{L}_{rec} + \lambda_{cls}\mathcal{L}_{cls},
\label{energyClassifier}
\end{equation}
where $\mathcal{L}_{cls}$ is cross entropy loss and $\lambda_{cls} = 0.0005$ to limit the gradient magnitude.
%will not be too large to deteriorate the BRDF reconstruction results. 

\subsection{Designing DCRFs for Refinement}
The prediction of our base network is quite reasonable. However, accuracy may further be enhanced by post-processing through a DCRF (trained end-to-end).
%there can be artifacts in some cases, thus, we design post-processing methods to enhance accuracy.

\vspace{-0.4cm}
\paragraph{Diffuse color refinement}
For diffuse prediction, when capturing the image of specular materials, parts of the surface might be saturated by specular highlight. This can sometimes lead to artifacts in the diffuse color prediction since the network has to hallucinate the diffuse color from nearby pixels. To remove such artifacts, we incorporate a densely connected continuous conditional random field (DCRF) \cite{continuousCRF} to smooth the diffuse color prediction. Let $\mathbf{\hat{d}}_{i}$ be the diffuse color prediction of network at pixel $i$, $\mathbf{p}_{i}$ be its position and $\mathbf{\bar{I}}_{i}$ is the normalized diffuse RGB color of the input image. We use the normalized color of the input image to remove the influence of light intensity when measuring the similarity between two pixels. The energy function of the dense connected CRF that is minimized over $\{\mathbf{d}_{i}\}$ for diffuse prediction is defined as: 
\begin{equation}
  \scriptsize
\!\!\!\! \sum_{i=1}^{N} \alpha_{i}^{d}(\mathbf{d}_{i} - \mathbf{\hat{d}}_{i})^{2} + \sum_{i, j}^{N}(\mathbf{d}_{i} - \mathbf{d}_{j})^{2} \!\! \left( \beta_{1}^{d}\kappa_{1}(\mathbf{p}_{i}; \mathbf{p}_{j})  
 + \beta_{2}^{d}\kappa_{2}(\mathbf{p}_{i}, \mathbf{\bar{I}}_{i}; \mathbf{p}_{j},\mathbf{\bar{I}}_{j}) + \beta_{3}^{d}\kappa_{3}(\mathbf{p}_{i}, \mathbf{\hat{d}}_{i}; \mathbf{p}_{j},\mathbf{\hat{d}}_{j}) \right).
\label{diffuseCRF}
\end{equation}
Here $\kappa_{i}$ are Gaussian smoothing kernels, while $\alpha_{i}^{d}$ and $\{\beta_{i}^{d}\}$ are coefficients to balance the contribution of unary and smoothness terms. Notice that we have a spatially varying $\alpha_{i}^{d}$ to allow different unary weights for different pixels. The intuition is that artifacts usually occur near the center of images with specular highlights. For those pixels, we should have lower unary weights so that the CRF learns to predict their diffuse color from nearby pixels.

\vspace{-0.4cm}
\paragraph{Normal refinement}
Once we have the refined diffuse color, we can use it to improve the prediction of other BRDF parameters. To reduce the noise in normal prediction, we use a DCRF with two smoothness kernels. One is based on the pixel position while the other is a bilateral kernel based on the position of the pixel and the gradient of the diffuse color. The intuition is that pixels with similar diffuse color gradients often have similar normal directions.
%\KS{but you are using gradient of diffuse color}\ZL{Correct the mistakes. Thanks!}. 
Let $\hat{\mathbf{n}}_{i}$ be the normal predicted by the network. The energy function for normal prediction is defined as
\begin{equation}
  \scriptsize
\min_{\{\mathbf{n}_{i}\}}: \sum_{i=1}^{N}\alpha^{n}(\mathbf{n}_{i} - \mathbf{\hat{n}}_{i})^{2} + \sum_{i, j}^{N}(\mathbf{n}_{i} - \mathbf{n}_{j})^{2}\Big(\beta_{1}^{n}\kappa_{1}(\mathbf{p}_{i};\mathbf{p}_{j}) 
+ \beta_{2}^{n}\kappa_{2}(\mathbf{p}_{i}, \Delta\mathbf{d}_{i}; \mathbf{p}_{j}, \Delta\mathbf{d}_{j}) \Big) 
\label{normalCRF}
\end{equation}

\vspace{-0.4cm}
\paragraph{Roughness refinement}
Since we use a collocated light source to illuminate the material, once we have the normal and diffuse color predictions, we can use them to estimate the roughness term by either grid search or using a gradient-based method. However, since the microfacet BRDF model is not convex nor monotonic with respect to the roughness term, there is no guarantee that we can find a global minimum. Also, due to noise from the normal and diffuse predictions, as well as environment lighting, it is difficult to get an accurate roughness prediction using optimization alone, especially when the glossiness in the image is not apparent. Therefore, we propose to combine the output of the network and the optimization method to get a more accurate roughness prediction. We use a DCRF with two unary terms, $\hat{r}_{i}$ and $\tilde{r}_{i}$, given by the network prediction and the coarse-to-fine grid search method of \cite{mobileBRDF}, respectively: 
\begin{equation}
  \scriptsize
\min_{ \{r_{i}\} }:  \sum_{i=1}^{N}\alpha_{i0}^{r}(r_{i} - \hat{r}_{i})^{2} + \alpha_{i1}^{r}(r_{i} - \tilde{r}_{i})^{2} + \sum_{i,j}^{N} (r_{i} - r_{j})^{2}
 \Big(\beta_{0}\kappa_{0}(\mathbf{p}_{i}; \mathbf{p}_{j}) + \beta_{1}\kappa_{1}(\mathbf{p}_{i}, \mathbf{d}_{i}; \mathbf{p}_{j}, \mathbf{d}_{j}) \Big)
\label{roughnessCRF} 
\end{equation}

All DCRF coefficients are learned in an end-to-end manner using \cite{learnCRF}. Here, we have a different set of DCRF parameters for each material type to increase model capacity. During both training and testing, the classifier output is used to average the parameters from different material types, to determine the DCRF parameters. More implementation details are in supplementary material. 

\vspace{-0.3cm}
\section{Experiments}
\label{sec:experiments}
\vspace{-0.2cm}

In this section, we demonstrate our method and compare it to baselines on a wide range of synthetic and real data. 

\vspace{-0.3cm}
\paragraph{Rendering synthetic training dataset}
To create our synthetic data, we apply the SVBRDFs on planar surfaces and render them using Mitsuba~\cite{Mitsuba} with the BRDF importance sampling suggested in \cite{brdfModel}. We choose a camera field of view of $43.35^{\circ}$ to mimic typical mobile phone cameras. 
%\KS{how do we set flash properties like falloff?}\ZL{Currently we assume a point light for illumination so the fall off will be Euclidean distance from the point light source.} 
To better model real-world lighting conditions, we render images under a combination of a dominant point light (flash) and an environment map. We use the $49$ environment maps used in \cite{CNN-BRDF}, with random rotations. We sample the light source position from a Gaussian distribution centered at the camera to make the inference robust to differences in real-world mobile phones. We render linear images, though clamped to $(0, 1)$ to mimic cameras with insufficient dynamic range. However, we still wish to reconstruct the full dynamic range of the SVBRDF parameters. To aid in this, we can render HDR images using in-our network rendering layer and compute reconstruction error w.r.t HDR ground truth images. In practice, this leads to unstable gradients in training; we mitigate this by applying a gamma of $2.2$ and minor clamping to $(0,1.5)$ when computing the image reconstruction loss. We find that this, in addition to our L2 losses on the SVBRDF parameters, allows us to hallucinate details from saturated images.     

\vspace{-0.3cm}
\paragraph{Training details}
We use Adam optimizer \cite{adam} to train our network. We set $\beta_{1} = 0.5$ when training the encoder and decoders and $\beta_{1} = 0.9$ when training the classifier. The initial learning rate is set to be $10^{-4}$ for the encoder, $2\times 10^{-4}$ for the three decoders and $2\times 10^{-5}$ for the classifier. We cut down the learning rate by half in every two epochs. Since we find that the diffuse color and normal direction contribute much more to the final appearance, we first train their encoder-decoders for 15 epochs, then we fix the encoder and train the roughness decoder separately for 8 epochs. 
%We train the network with 15 epochs when doing diffuse and normal prediction, 8 epochs for roughness prediction.
Next, we fix the network and train the parameters for the DCRFs, using Adam optimizer to update their coefficients.

%\vspace{-0.4cm}
%\paragraph{Capturing real dataset}
%We capture real data with Huawei P9 mobile phone. The images are of size $3968\times 2976$. We calibrate the camera of the phone and crop a central patch of size $2392\times 2392$, to obtain a $43.35^{\circ}$ field of view. We also calibrate the position of the flash light by taking an image of a white planar surface and finding the pixel with the largest intensity as the center of flash light.

\vspace{-0.2cm}
\subsection{Results on Synthetic Data}

\begin{figure}[!!t]
\centering
\setlength{\tabcolsep}{1pt}
\begin{tabular}{cccccc@{\hspace{0.2cm}}cccccc}
\centering
& \scriptsize{Input} & \scriptsize{Albedo} & \scriptsize{Normals} & \scriptsize{Roughness} & \scriptsize{Relit}& & \scriptsize{Input} & \scriptsize{Albedo} & \scriptsize{Normals} & \scriptsize{Roughness} & \scriptsize{Relit} \\
\rotatebox[origin=c]{90}{\scriptsize{\qquad\qquad GT}} & 
\includegraphics[width=0.40in]{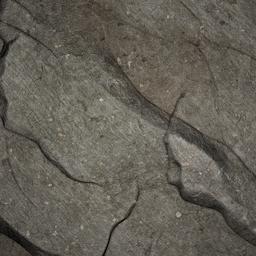} & 
\includegraphics[width=0.40in]{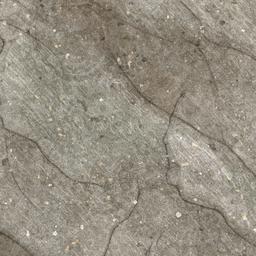} & 
\includegraphics[width=0.40in]{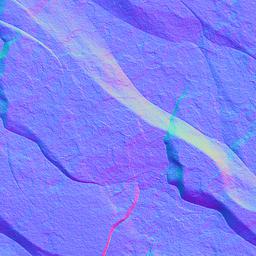} & 
\includegraphics[width=0.40in]{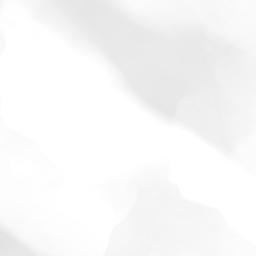} & 
\includegraphics[width=0.40in]{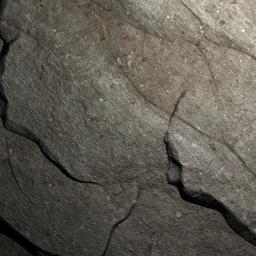} & 
\rotatebox[origin=c]{90}{\scriptsize{\qquad\qquad GT}} & 
\includegraphics[width=0.40in]{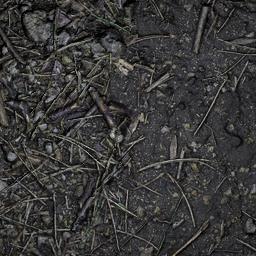} & 
\includegraphics[width=0.40in]{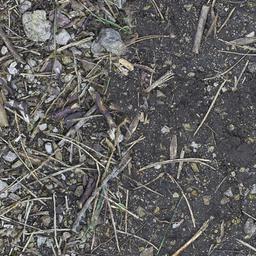} & 
\includegraphics[width=0.40in]{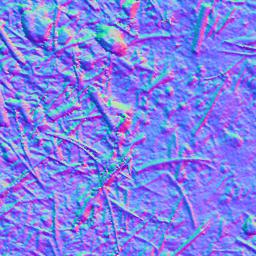} & 
\includegraphics[width=0.40in]{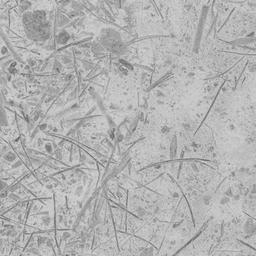} & 
\includegraphics[width=0.40in]{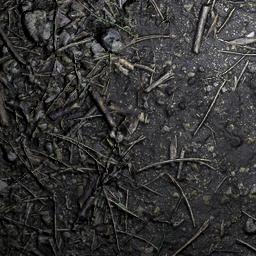} 
\vspace{-0.5cm} \\ 
\rotatebox[origin=c]{90}{\scriptsize{\qquad\qquad Pred}}  & 
\includegraphics[width=0.40in]{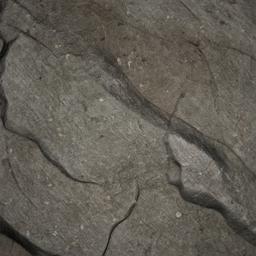} & 
\includegraphics[width=0.40in]{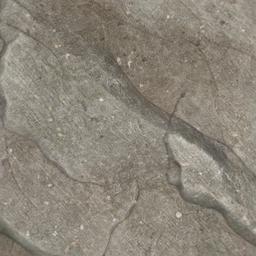} & 
\includegraphics[width=0.40in]{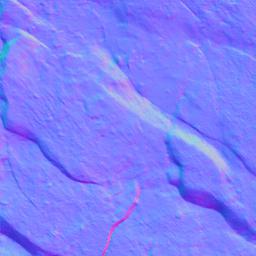} & 
\includegraphics[width=0.40in]{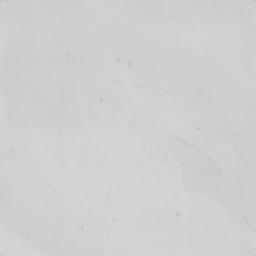} & 
\includegraphics[width=0.40in]{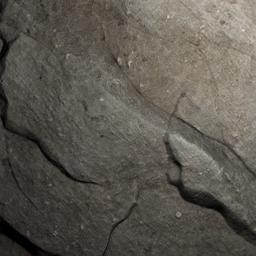} &
\rotatebox[origin=c]{90}{\scriptsize{\qquad\qquad Pred}} &
\includegraphics[width=0.40in]{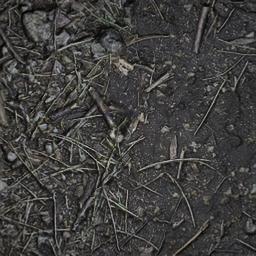} & 
\includegraphics[width=0.40in]{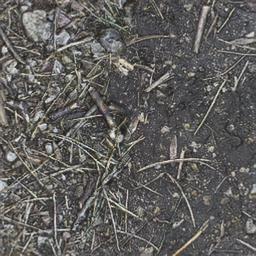} & 
\includegraphics[width=0.40in]{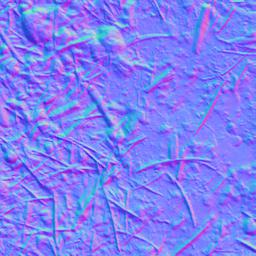} & 
\includegraphics[width=0.40in]{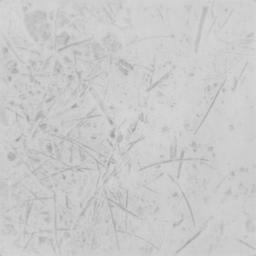} & 
\includegraphics[width=0.40in]{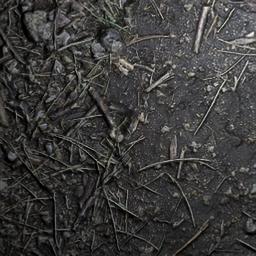}
\vspace{-0.5cm}\\ 
\rotatebox[origin=c]{90}{\scriptsize{\qquad\qquad GT}} & 
\includegraphics[width=0.40in]{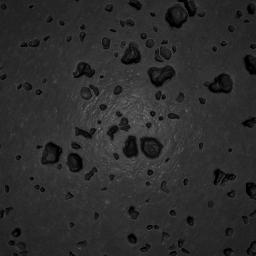} & 
\includegraphics[width=0.40in]{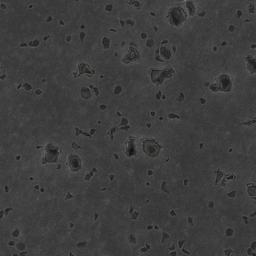} & 
\includegraphics[width=0.40in]{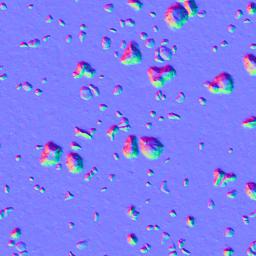} & 
\includegraphics[width=0.40in]{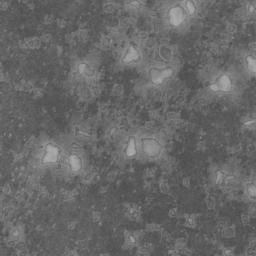} & 
\includegraphics[width=0.40in]{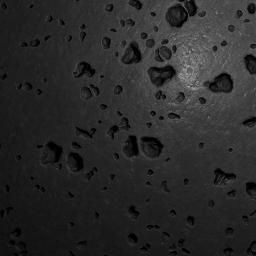} & 
\rotatebox[origin=c]{90}{\scriptsize{\qquad\qquad GT}} & 
\includegraphics[width=0.40in]{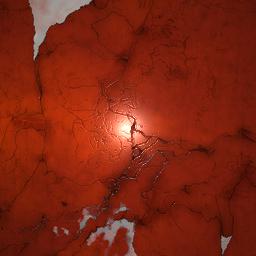} & 
\includegraphics[width=0.40in]{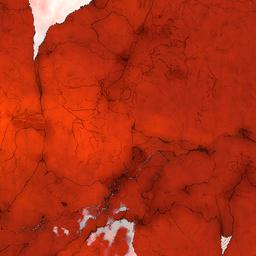} & 
\includegraphics[width=0.40in]{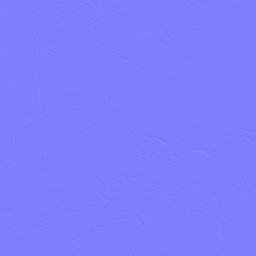} & 
\includegraphics[width=0.40in]{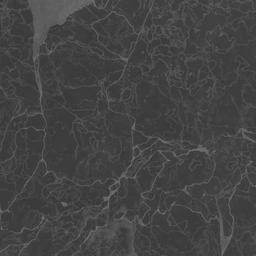} & 
\includegraphics[width=0.40in]{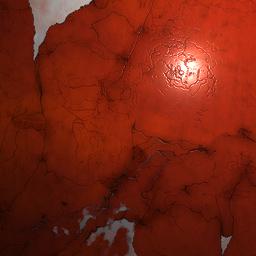} 
\vspace{-0.5cm}\\ 
\rotatebox[origin=c]{90}{\scriptsize{\qquad\qquad Pred}}  & 
\includegraphics[width=0.40in]{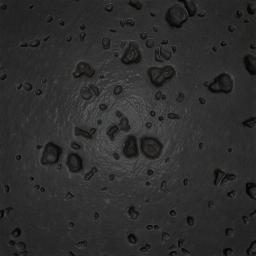} & 
\includegraphics[width=0.40in]{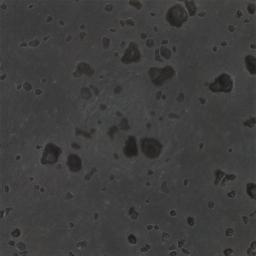} & 
\includegraphics[width=0.40in]{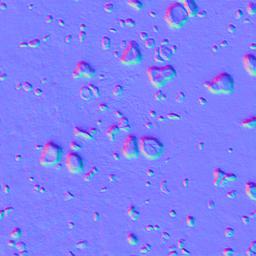} & 
\includegraphics[width=0.40in]{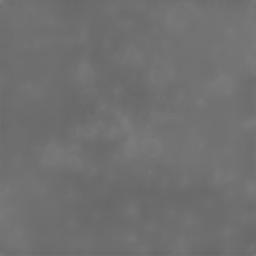} & 
\includegraphics[width=0.40in]{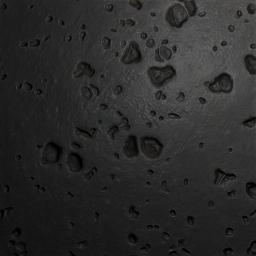} &
\rotatebox[origin=c]{90}{\scriptsize{\qquad\qquad Pred}} &
\includegraphics[width=0.40in]{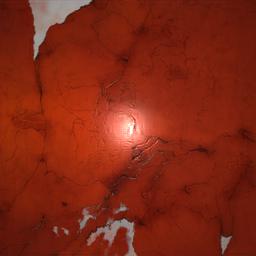} & 
\includegraphics[width=0.40in]{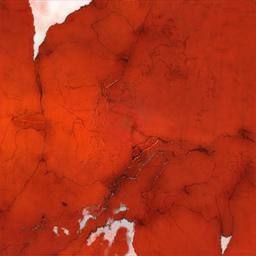} & 
\includegraphics[width=0.40in]{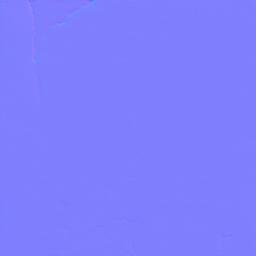} & 
\includegraphics[width=0.40in]{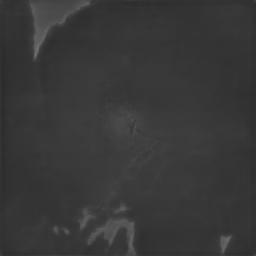} & 
\includegraphics[width=0.40in]{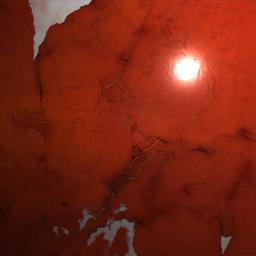}
\vspace{-0.5cm}\\ 
\rotatebox[origin=c]{90}{\scriptsize{\qquad\qquad GT}} & 
\includegraphics[width=0.40in]{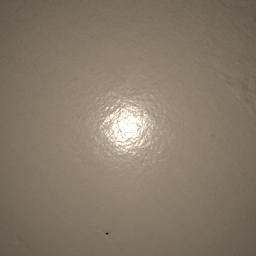} & 
\includegraphics[width=0.40in]{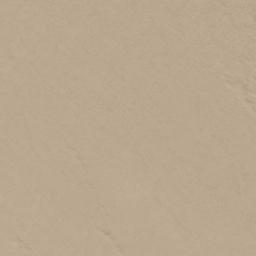} & 
\includegraphics[width=0.40in]{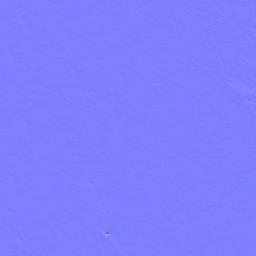} & 
\includegraphics[width=0.40in]{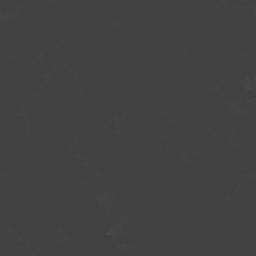} & 
\includegraphics[width=0.40in]{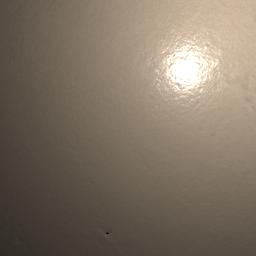} & 
\rotatebox[origin=c]{90}{\scriptsize{\qquad\qquad GT}} & 
\includegraphics[width=0.40in]{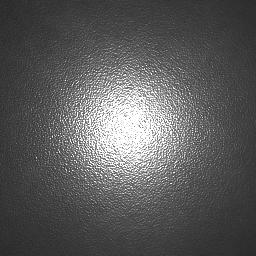} & 
\includegraphics[width=0.40in]{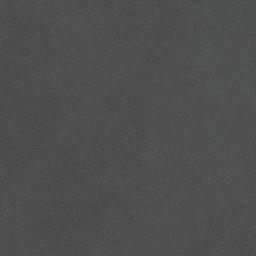} & 
\includegraphics[width=0.40in]{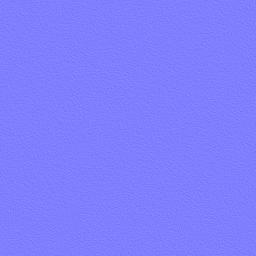} & 
\includegraphics[width=0.40in]{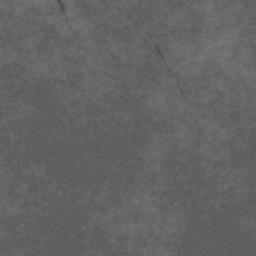} & 
\includegraphics[width=0.40in]{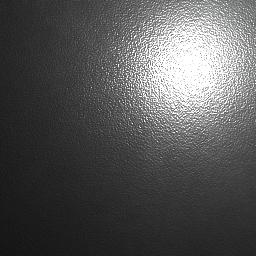} 
\vspace{-0.5cm}\\ 
\rotatebox[origin=c]{90}{\scriptsize{\qquad\qquad Pred}}  & 
\includegraphics[width=0.40in]{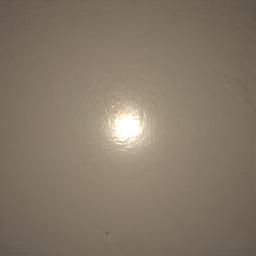} & 
\includegraphics[width=0.40in]{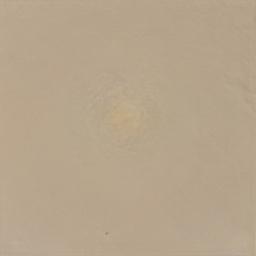} & 
\includegraphics[width=0.40in]{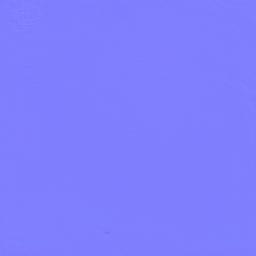} & 
\includegraphics[width=0.40in]{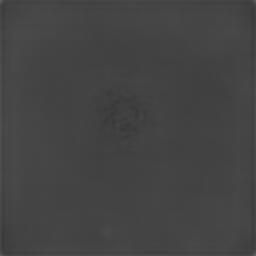} & 
\includegraphics[width=0.40in]{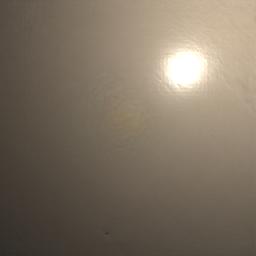} &
\rotatebox[origin=c]{90}{\scriptsize{\qquad\qquad Pred}} &
\includegraphics[width=0.40in]{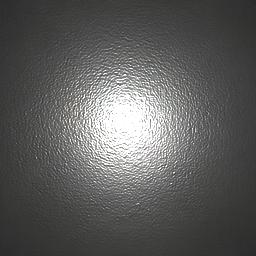} & 
\includegraphics[width=0.40in]{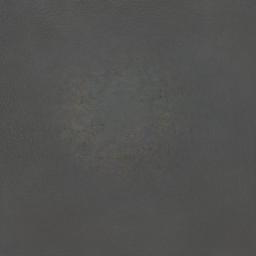} & 
\includegraphics[width=0.40in]{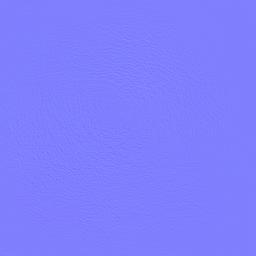} & 
\includegraphics[width=0.40in]{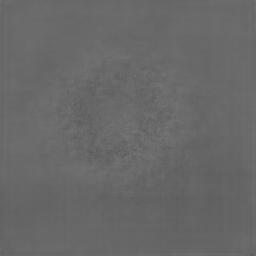} & 
\includegraphics[width=0.40in]{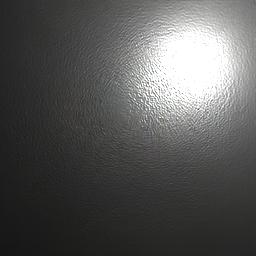}
\vspace{-0.5cm}
\end{tabular}
\label{fig:brdfrecon}
\vspace{-0.5cm}
\caption{BRDF reconstruction results from our full method ($\mathtt{clsCRF}$-$\mathtt{pt}$ in Table \ref{quantitativeSyn}) on the test set. We compare the ground truth parameters with our reconstructions as well as renderings of these parameters under novel lighting. The accuracy of our renderings indicates the accuracy of our method.}
\vspace{-0.3cm}
\end{figure}

\begin{figure}[!!t]
\centering
\begin{tabular}{cccccc|ccccc}
 & \scriptsize{Env1} & \scriptsize{Env2} & \scriptsize{$\phi=180^{\circ}$} & \scriptsize{$\phi=90^{\circ}$}& \scriptsize{$\phi = 270^{\circ}$} & \scriptsize{Env1} & \scriptsize{Env2} & \scriptsize{$\phi=180^{\circ}$} & \scriptsize{$\phi=90^{\circ}$}& \scriptsize{$\phi = 270^{\circ}$} \\
\raisebox{0.3cm}{\rotatebox[origin=c]{90}{\small{Pred}}} &
\includegraphics[width=0.40in]{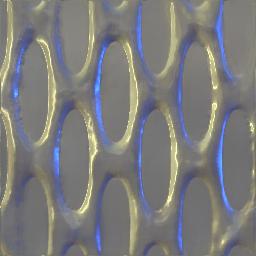} &
\includegraphics[width=0.40in]{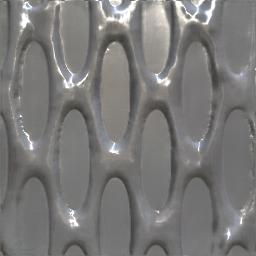} &
\includegraphics[width=0.40in]{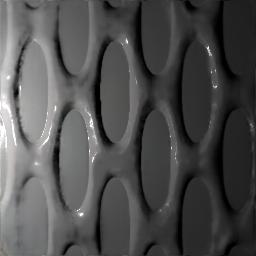} &
\includegraphics[width=0.40in]{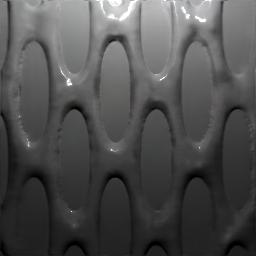} &
\includegraphics[width=0.40in]{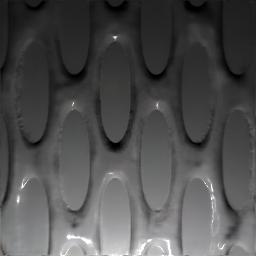} &
\includegraphics[width=0.40in]{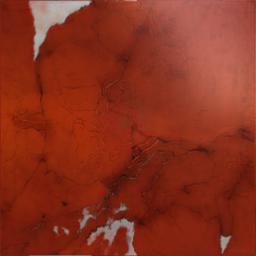} &
\includegraphics[width=0.40in]{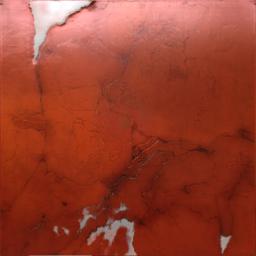} &
\includegraphics[width=0.40in]{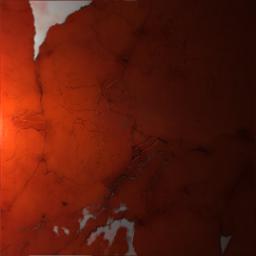} &
\includegraphics[width=0.40in]{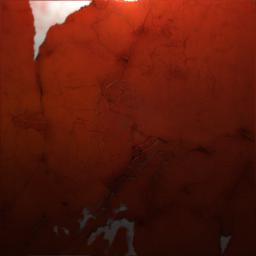} &
\includegraphics[width=0.40in]{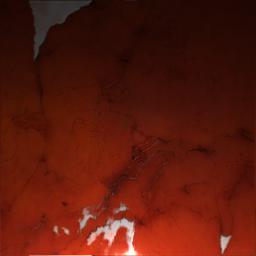} 
\\
\raisebox{0.3cm}{\rotatebox[origin=c]{90}{\small{GT}}} &
\includegraphics[width=0.40in]{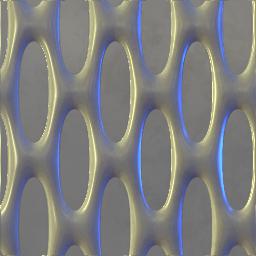} &
\includegraphics[width=0.40in]{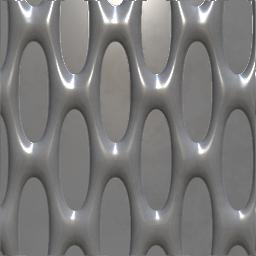} &
\includegraphics[width=0.40in]{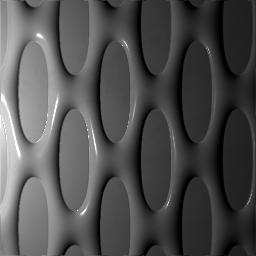} &
\includegraphics[width=0.40in]{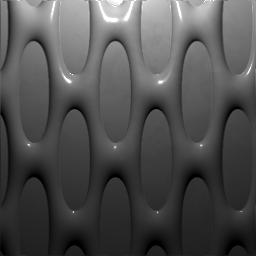} &
\includegraphics[width=0.40in]{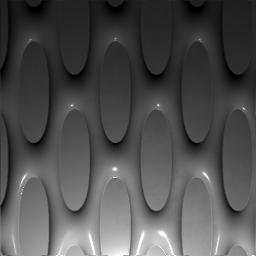} &
\includegraphics[width=0.40in]{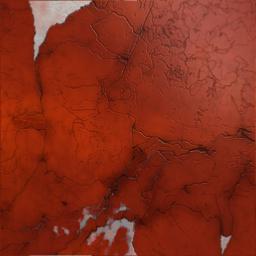} &
\includegraphics[width=0.40in]{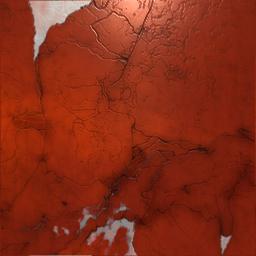} &
\includegraphics[width=0.40in]{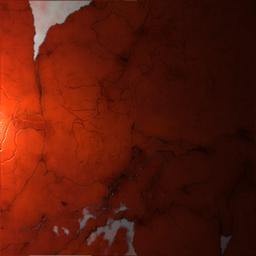} &
\includegraphics[width=0.40in]{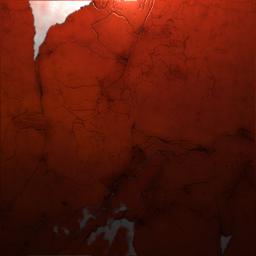} &
\includegraphics[width=0.40in]{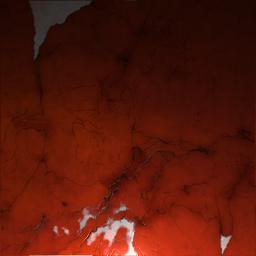} 
\end{tabular}
\vspace{-0.3cm}
\caption{\footnotesize Materials estimated with our method and rendered under two environment lights and three point lights (placed on a unit sphere at $\theta = 50^{\circ}$ and various $\phi$ angles).}
\label{syncRelighting}
\vspace{-0.4cm}
\end{figure}

\vspace{-0.2cm}
\paragraph{Qualitative results}
Figure \ref{fig:brdfrecon} shows results of our network on our synthetic test dataset. We can observe that spatially varying surface normals, diffuse albedo and roughness are recovered at high quality, which allows relighting under novel light source directions that are very different from the input. To further demonstrate our BRDF reconstruction quality, in Figure \ref{syncRelighting}, we show relighting results under different environment maps and point lights at oblique angles.
%When rendered with point light sources, we set the angle between the light and the $z$-axis to be $50^{\circ}$.
%In the main paper and in Figure \ref{errorExample}, this angle is set to be $30^{\circ}$. From Figure \ref{errorExample},
Note that our relighting results closely match the ground truth even under different lighting conditions; this indicates the accuracy of our reconstructions.

We next perform quantitative ablation studies to evaluate various components of our network design and study comparisons to prior work.

\begin{figure}[!!t]
  \vspace{-0.4cm}
\begin{minipage}[c]{0.5\textwidth}
\centering
\begin{tabular}{cccccc}
& \scriptsize{\texttt{cls-env}} & \scriptsize{\texttt{basic-pt}} & \scriptsize{\texttt{cls-pt}} & \scriptsize{\texttt{clsCRF-pt}} & \scriptsize{\texttt{GT}} \\
\raisebox{0.5cm}{\rotatebox[origin=c]{90}{\scriptsize{Albedo}}} &
\includegraphics[width=0.45in]{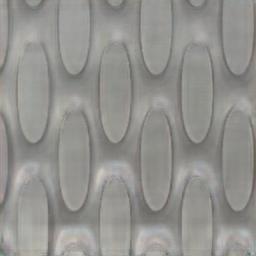} &
\includegraphics[width=0.45in]{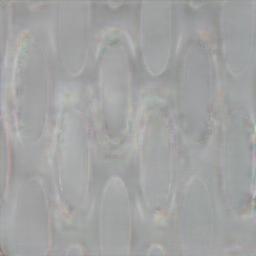} &
\includegraphics[width=0.45in]{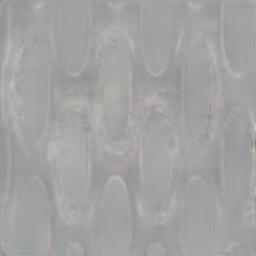} & 
\includegraphics[width=0.45in]{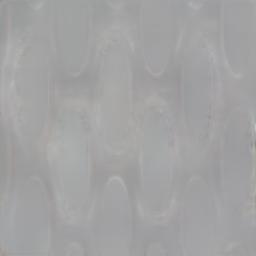} &
\includegraphics[width=0.45in]{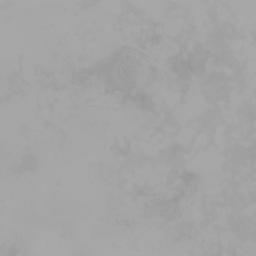} \\[-0.0cm]
%\vspace{-1.0cm}\\
\raisebox{0.5cm}{\rotatebox[origin=c]{90}{\scriptsize{Normal}}} &
\includegraphics[width=0.45in]{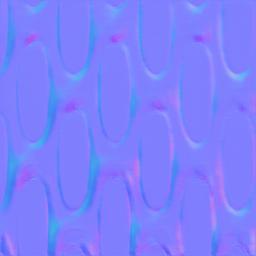} &
\includegraphics[width=0.45in]{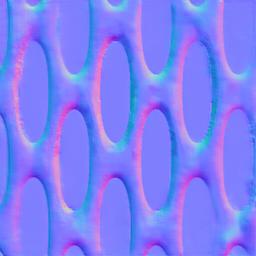} &
\includegraphics[width=0.45in]{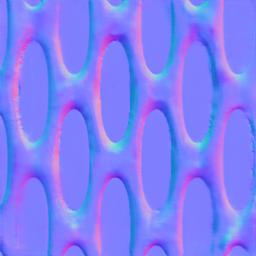} & 
\includegraphics[width=0.45in]{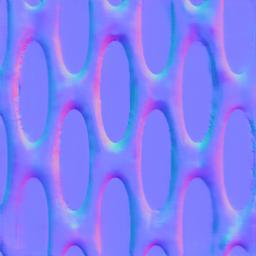} &
\includegraphics[width=0.45in]{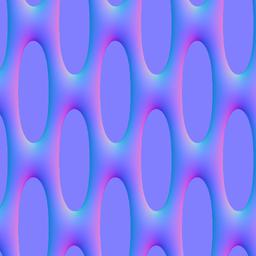} \\[-0.05cm]
%\vspace{-1.0cm}\\
\raisebox{0.5cm}{\rotatebox[origin=c]{90}{\scriptsize{Roughness}}} &
\includegraphics[width=0.45in]{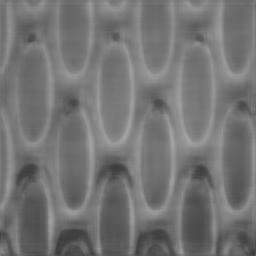} &
\includegraphics[width=0.45in]{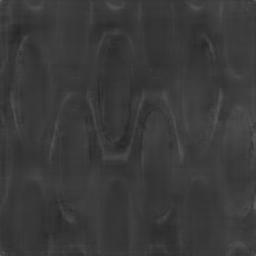} &
\includegraphics[width=0.45in]{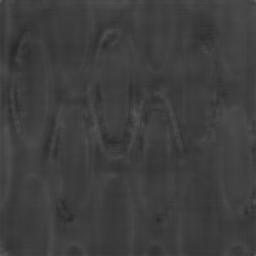} & 
\includegraphics[width=0.45in]{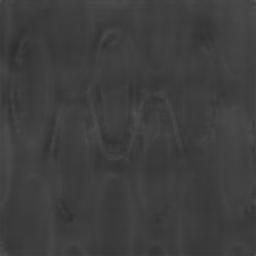} &
\includegraphics[width=0.45in]{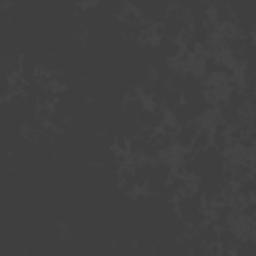} \\[-0.0cm]
%\vspace{-1.2cm}\\
\raisebox{0.4cm}{\rotatebox[origin=c]{90}{\scriptsize{Relight}}} &
\includegraphics[width=0.45in]{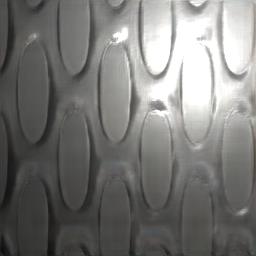} &
\includegraphics[width=0.45in]{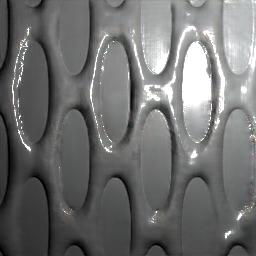} &
\includegraphics[width=0.45in]{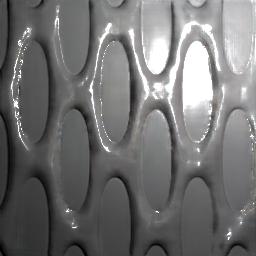} & 
\includegraphics[width=0.45in]{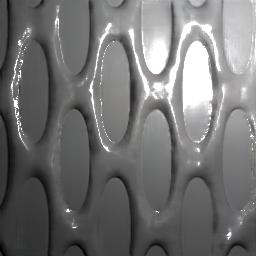} &
\includegraphics[width=0.45in]{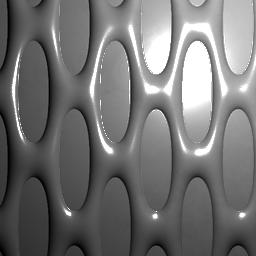}
%\vspace{-1.2cm}\\
\end{tabular}
%\vspace{-0.1cm}
\caption{Qualitative comparison of BRDF reconstruction results of different variants of our network. The notation is the same as Table~\ref{quantitativeSyn} and $-\mathtt{env}$ represents environment illumination.}
\label{classifier}
\end{minipage}\hfill
\begin{minipage}[c]{0.45\textwidth}
\small    
\begin{tabular}{|c|l|ccc|}
\hline
& & Albedo-N & Normals & Rough \\
& & ($e^{-4}$) & ($e^{-3}$) & ($e^{-2}$) \\
\hline
\multirow{4}{*}{\rotatebox[origin=c]{90}{\protect\cite{CNN-BRDF} }} 
& metal &91.8 &27.2 & --  \\
& wood &35.9 &11.2 & -- \\
& plastic &12.5 &17.6 & --  \\
& Total &56.1 &19.7 & -- \\
\hline
\multirow{4}{*}{\rotatebox[origin=c]{90}{$\mathtt{cls}$-$\mathtt{env}$} } 
& metal &54.9 &25.2 &13.4  \\
& wood &13.7 &11.1 &19.5  \\
& plastic &7.96 &14.2 &25.3  \\
& Total &30.9 &18.1 &18.0  \\
\hline
\multirow{4}{*}{\rotatebox[origin=c]{90}{$\mathtt{cls}$-$\mathtt{pt}$} } 
&metal 		&21.7 	&15.1 	&4.06  \\
&wood  		&3.53 	&8.75 	&4.40  \\
&plastic 	&1.64 	&9.10 	&7.24  \\
&Total 		&11.3 	&11.7 	&4.83\\
\hline
\end{tabular}
%\vspace{-0.1cm}
\captionof{table}{BRDF reconstruction accuracy for different material types in our test set. Albedo-N is normalized diffuse albedo as in \cite{CNN-BRDF}, that is, the average norm of each pixel will be $0.5$.}
\label{quantitativeType}
\end{minipage}\hfill
%\vspace{-0.4cm}
\end{figure}

\vspace{-0.3cm}
\paragraph{Effects of material classifier and DCRF:}
The ablation study summarized in Table \ref{quantitativeSyn} shows that adding the material classifier reduces the L2 error for SVBRDF and normal estimation, as well as rendering error. This validates the intuition that the network can exploit the correlation between BRDF parameters and material type to produce better estimates.
We also observe that training the classifier together with the BRDF reconstruction network results in a material classification error of $73.65\%$, which significantly improves over just our pure material classification network that achieves $54.96\%$. This indicates that features trained for BRDF estimation are also useful for material recognition. In our experiments, incorporating the classifier without using its output to fuse BRDF reconstruction results does not improve BRDF estimation.
Figure \ref{classifier} shows the reconstruction result on a sample where the classifier and the DCRF qualitatively improve the BRDF estimation, especially for the diffuse albedo.

\begin{table}[!!t]
\begin{minipage}[c]{0.6\textwidth}
\small
\centering
\begin{tabular}{l|cccc}
\hline
Method & $\mathtt{basic}$-$\mathtt{pt}$ & $\mathtt{cls}$-$\mathtt{pt}$ & $\mathtt{cls}$$\mathtt{CRF}$-$\mathtt{pt}$ & $\mathtt{clsOnly}$-$\mathtt{pt}$ \\
\hline
Albedo ($e^{-3}$) & 7.78 & 7.58 & \textbf{7.42} & \\
Normal ($e^{-2}$) & 1.55 & 1.52 & \textbf{1.50} & \\
Rough ($e^{-2}$) & 8.75 & 8.55 & \textbf{8.53} & \\
%Render ($e^{-3}$) & 5.28 &	\textbf{5.16} 	& 5.22 & \\
Classify (\%) &	 & \textbf{73.65} & \textbf{73.65} & 54.96 \\
\hline
\end{tabular}
\end{minipage}\hfill
\begin{minipage}[c]{0.35\textwidth}
\caption{Left to right: basic encoder-decoder, adding material classifier, adding DCRF and a pure material classifier. $-\mathtt{pt}$ indicates training and testing with dominant point and environment lighting.}
%\caption{Network design ablation study. $\mathtt{basic}$ is our basic encoder-decoder network architecture, $\mathtt{cls}$ adds the material classfier, and $\mathtt{clsCRF}$ further adds the DCRF. $\mathtt{clsOnly}$ is a pure material classification network. $-\mathtt{pt}$ indicates the network was trained and tested on images with dominant point and environment illumination.}
\label{quantitativeSyn}
\end{minipage}\hfill
\end{table}
\vspace{-0.2cm}

\begin{figure}[!!t]
  \vspace{-0.6cm}
\centering
\begin{minipage}[c]{0.50\textwidth}
\raisebox{-2.2cm}{
\begin{tabular}{cccc}
\includegraphics[width=0.45in]{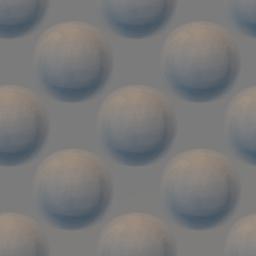} &
\includegraphics[width=0.45in]{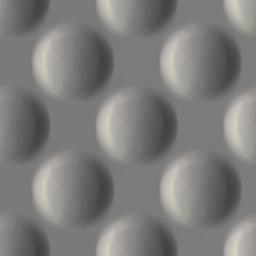} & 
\includegraphics[width=0.45in]{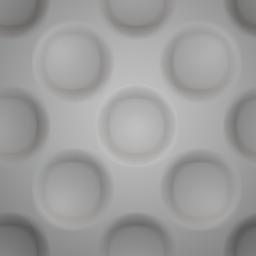} & 
\includegraphics[width=0.45in]{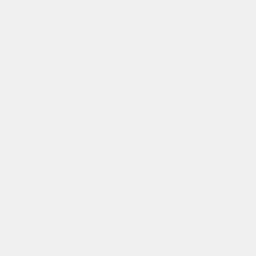} \\
{\scriptsize Input 1 \cite{CNN-BRDF}} & {\scriptsize Input 2 \cite{CNN-BRDF}} & {\scriptsize Our Input} & {\scriptsize Diffuse GT} \\
\includegraphics[width=0.45in]{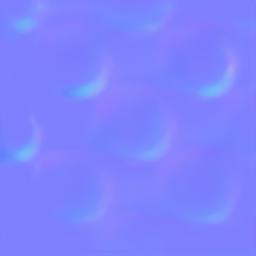} &
\includegraphics[width=0.45in]{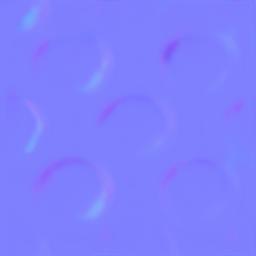} & 
\includegraphics[width=0.45in]{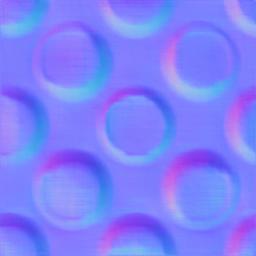} & 
\includegraphics[width=0.45in]{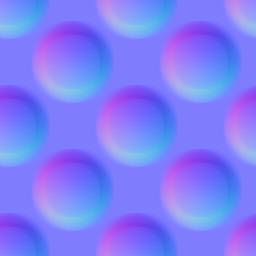} \\
{\scriptsize Normal 1 \cite{CNN-BRDF}} & {\scriptsize Normal 2 \cite{CNN-BRDF}} & {\scriptsize Our Normals} & {\scriptsize Normal GT} 
\end{tabular}
}
\end{minipage}\hfill
\begin{minipage}[c]{0.40\textwidth}
\caption{\small
  The first two inputs rendered under different environment maps are very different. Thus, the normals recovered using~\cite{CNN-BRDF} are inaccurate. Our method uses point illumination (third input) which alleviates the problem, and produces better normals.}
\label{pointVSenv}
\end{minipage}
\vspace{-0.4cm}
\end{figure}

\vspace{-0.2cm}
\paragraph{Effect of acquisition under point illumination}
Next we evaluate the effect of using point illumination during acquisition. For this, we train and test two variants of our full network -- one on images rendered under only environment illumination (-$\mathtt{env}$) and another on images illuminated by a point light besides environment illumination (-$\mathtt{pt}$).
%All rendered test images are normalized to an average energy of $0.5$. 
Results are in Table~\ref{quantitativeType} with qualitative visualizations in Figure ~\ref{classifier}. The model from \cite{CNN-BRDF} in Table~\ref{quantitativeType}, which is trained for environment lighting, performs slightly worse than our environment lighting network \texttt{cls-env}. But our network trained and evaluated on point and environment lighting, \texttt{cls-pt}, easily outperforms both. We argue this is because a collocated point light creates more consistent illumination across training and test images, while also capturing higher frequency information. Figure \ref{pointVSenv} illustrates this: the appearance of the same material under different environment lighting can significantly vary and the network has to be invariant to this, limiting reconstruction quality.

%KS{I say we drop this figure. We are saying our results are better than [20], does this mean we are learning a more invariant representation? Also we don't show comparison of our \texttt{cls-env} method which might be a nicer comparison. Also we need the space.} \ZL{Originally we add this figure so we can show intuitively how difficult it is to estimate normal under totally unconstrained illumination condition such that for some difficult examples, even human can not tell the whether the shape is convex or concave.}

\begin{figure}[!!t]
\centering
\begin{minipage}[c]{0.70\textwidth}
\raisebox{-1.5cm}{
\begin{tabular}{cccc}
\includegraphics[width=0.8in]{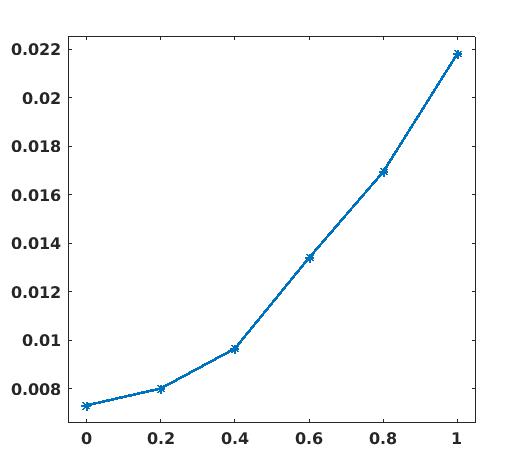} & 
\includegraphics[width=0.8in]{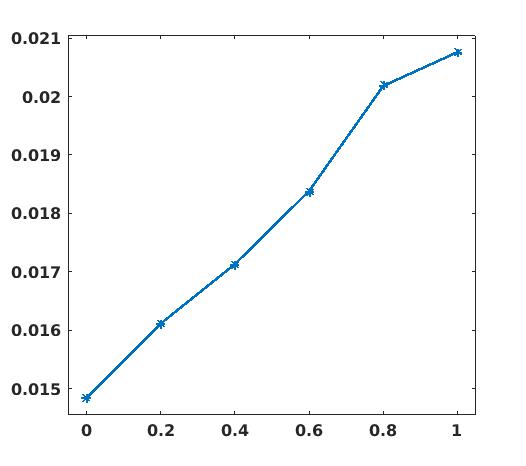} &
\includegraphics[width=0.8in]{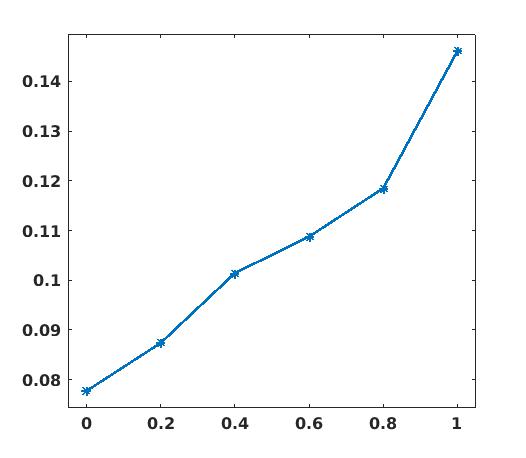} & 
\includegraphics[width=0.8in]{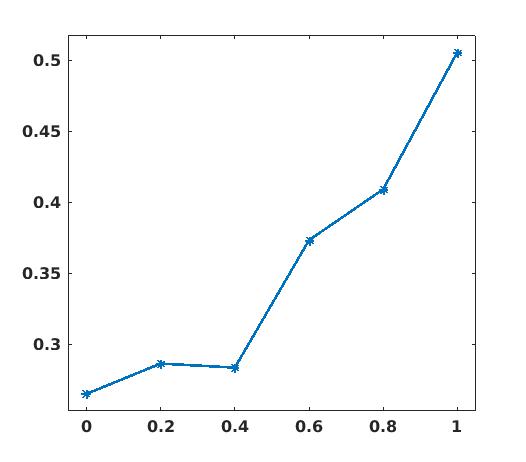} \\[-0.1cm]
\scriptsize{Albedo} & \scriptsize{Normal} & \scriptsize{Roughness} & \scriptsize{Classification}
\end{tabular}
}
\end{minipage}\hfill
\begin{minipage}[c]{0.28\textwidth}
\caption{\footnotesize SVBRDF estimation errors for relative intensities of environment against point light ranging from $0$ to $0.8$.}
\label{pointProportion}
\end{minipage}\hfill
\vspace{-0.5cm}
\end{figure}

\vspace{-0.2cm}
\paragraph{Relative effects of flash and environment light intensities}
In Figure \ref{pointProportion}, we train and test on a range of relative flash intensities, showing our network works well for each. Note that as relative flash intensity decreases, errors increase, which justifies our use of flash light. Using flash and no-flash pairs can help remove environment lighting, but needs alignment of two images, which limits applicability. 
%\KS{Can we mark [20] on these figures as a comparitive baseline?}\ZL{This might be unnecessary, since [20] (now [10]) train and test their network for specific kinds of materials while we train and test on the whole dataset.}

\vspace{-0.2cm}
\subsection{Results on Real Data}

%\KS{move para about real data acquisition here. Talk about how we capture, how we pre-process (i.e., tone-mapping) and then present results.}

\vspace{-0.3cm}
\paragraph{Acquisition setup}
To verify the generalizabity of our method to real data, we show results on real images captured with different mobile devices in both indoor and outdoor environments. We capture linear RAW  images (with potentially clipped highlights) with the flash enabled, using the Adobe Lightroom Mobile app. The mobile phones were hand-held and the optical axis of the camera was only approximately perpendicular to the surfaces (see Figure~\ref{teaser}).%We crop a central patch of size $2392\times 2392$, to obtain an approximately $43.35^{\circ}$ field of view. 

\begin{figure}[!!t]
\centering
\setlength{\tabcolsep}{1pt}
\begin{tabular}{ccccc @{\hspace{0.2cm}} ccccc}
\centering
 \scriptsize{Input} & \scriptsize{Albedo} & \scriptsize{Normals} & \scriptsize{Roughness} & \scriptsize{Rendering} & \scriptsize{Input} & \scriptsize{Albedo} & \scriptsize{Normals} & \scriptsize{Roughness} & \scriptsize{Rendering} \\
\includegraphics[width=0.44in]{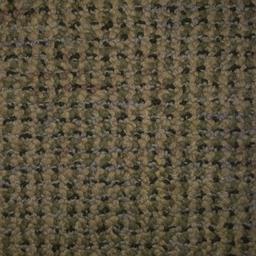} & 
\includegraphics[width=0.44in]{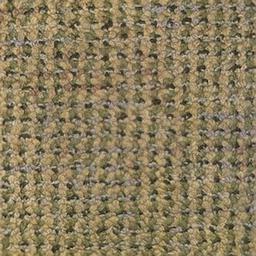} & 
\includegraphics[width=0.44in]{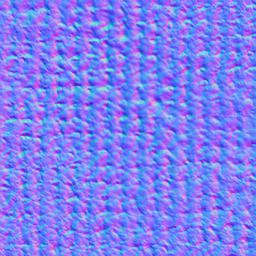} & 
\includegraphics[width=0.44in]{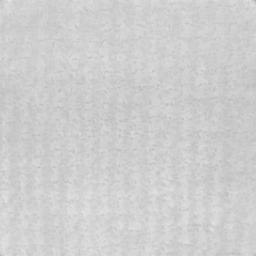} & 
\includegraphics[width=0.44in]{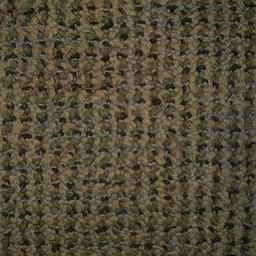} & 
\includegraphics[width=0.44in]{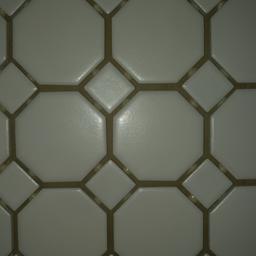} & 
\includegraphics[width=0.44in]{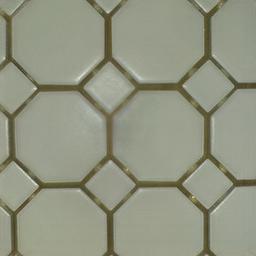} & 
\includegraphics[width=0.44in]{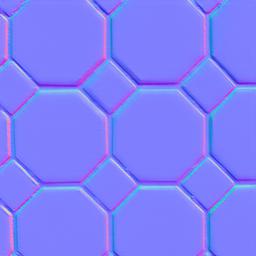} & 
\includegraphics[width=0.44in]{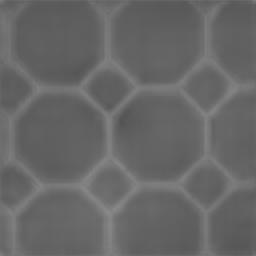} & 
\includegraphics[width=0.44in]{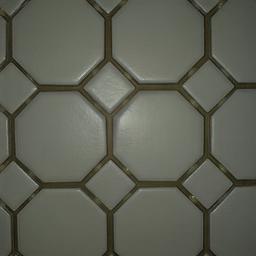} 
\vspace{0cm}\\ 
\includegraphics[width=0.44in]{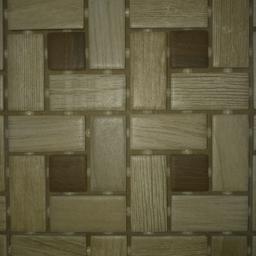} & 
\includegraphics[width=0.44in]{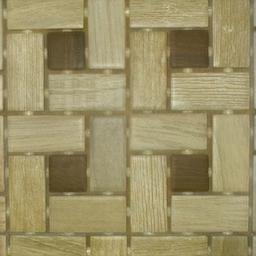} & 
\includegraphics[width=0.44in]{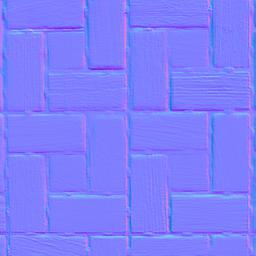} & 
\includegraphics[width=0.44in]{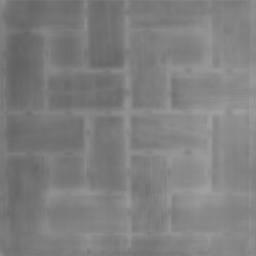} & 
\includegraphics[width=0.44in]{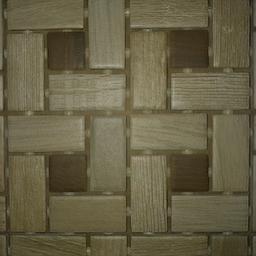} & 
\includegraphics[width=0.44in]{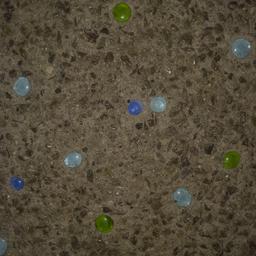} & 
\includegraphics[width=0.44in]{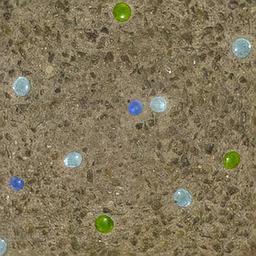} & 
\includegraphics[width=0.44in]{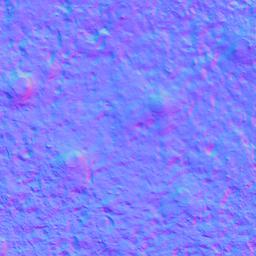} & 
\includegraphics[width=0.44in]{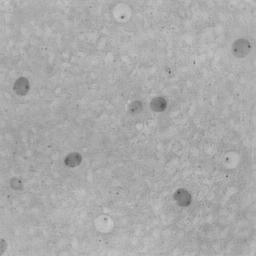} & 
\includegraphics[width=0.44in]{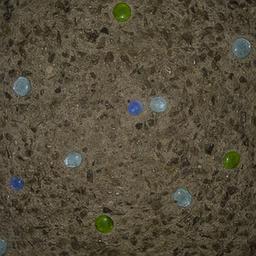} 
\vspace{0cm}\\ 
\includegraphics[width=0.44in]{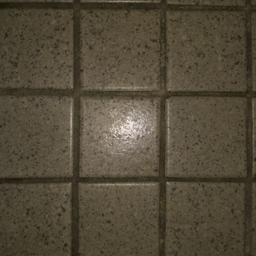} & 
\includegraphics[width=0.44in]{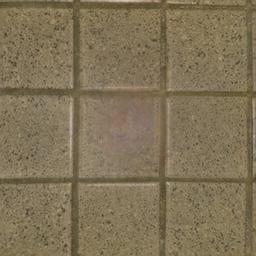} & 
\includegraphics[width=0.44in]{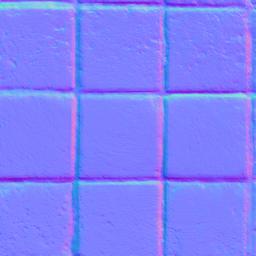} & 
\includegraphics[width=0.44in]{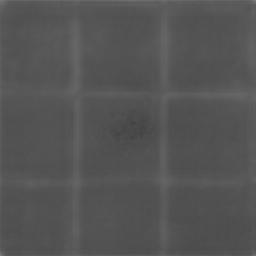} & 
\includegraphics[width=0.44in]{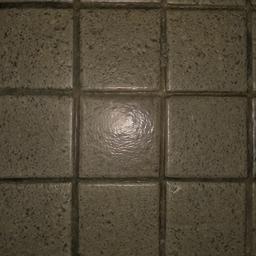} & 
\includegraphics[width=0.44in]{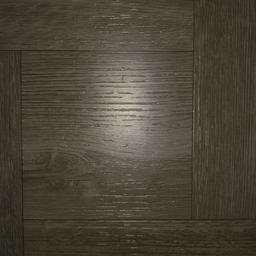} & 
\includegraphics[width=0.44in]{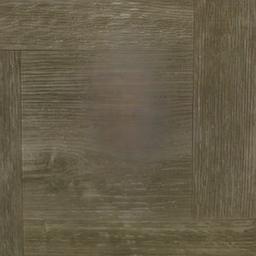} & 
\includegraphics[width=0.44in]{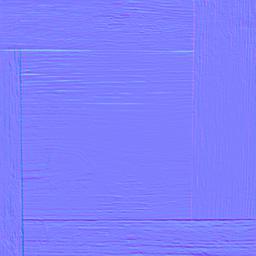} & 
\includegraphics[width=0.44in]{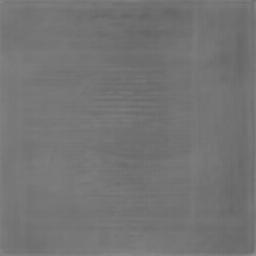} & 
\includegraphics[width=0.44in]{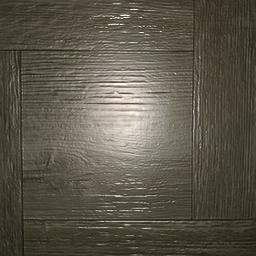} 
\vspace{0cm}\\
\includegraphics[width=0.44in]{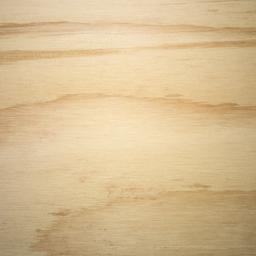} & 
\includegraphics[width=0.44in]{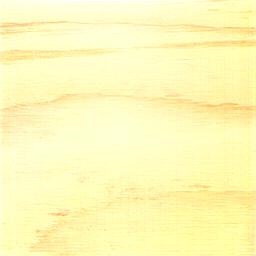} & 
\includegraphics[width=0.44in]{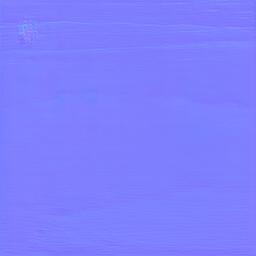} & 
\includegraphics[width=0.44in]{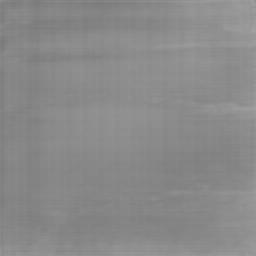} & 
\includegraphics[width=0.44in]{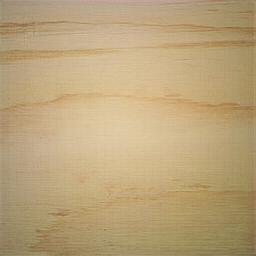} &
\includegraphics[width=0.44in]{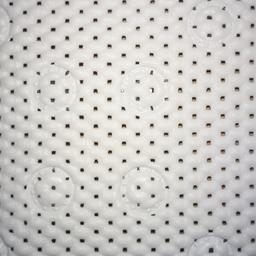} & 
\includegraphics[width=0.44in]{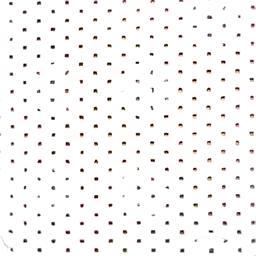} & 
\includegraphics[width=0.44in]{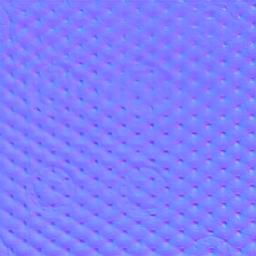} & 
\includegraphics[width=0.44in]{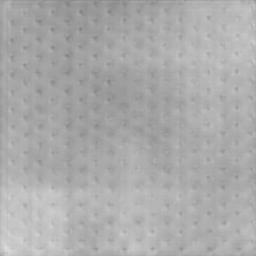} & 
\includegraphics[width=0.44in]{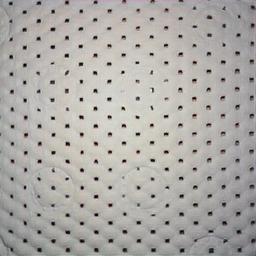} 
\vspace{0cm}\\
\includegraphics[width=0.44in]{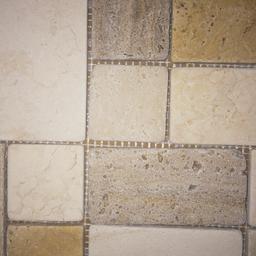} & 
\includegraphics[width=0.44in]{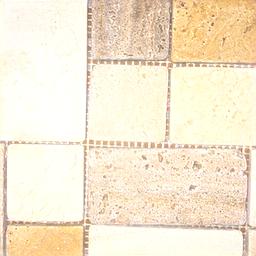} & 
\includegraphics[width=0.44in]{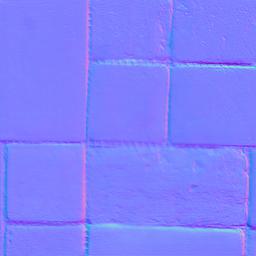} & 
\includegraphics[width=0.44in]{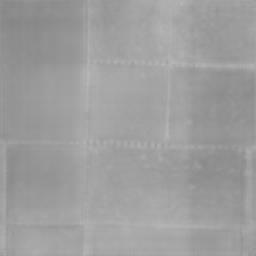} & 
\includegraphics[width=0.44in]{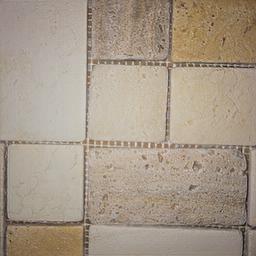} &
\includegraphics[width=0.44in]{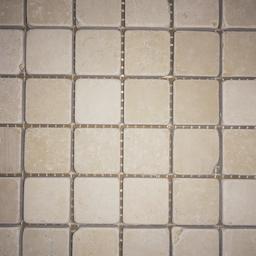} & 
\includegraphics[width=0.44in]{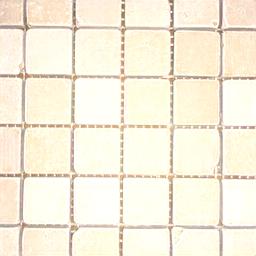} & 
\includegraphics[width=0.44in]{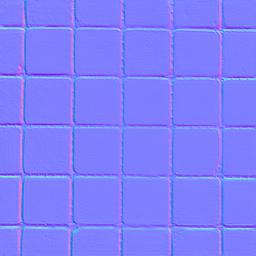} & 
\includegraphics[width=0.44in]{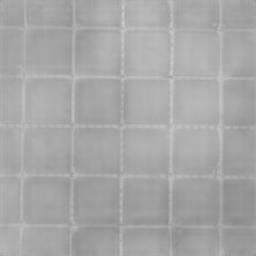} & 
\includegraphics[width=0.44in]{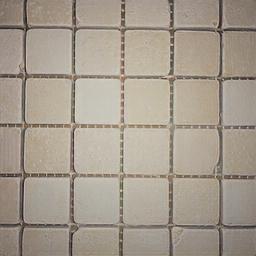} 
\end{tabular}
\vspace{-0.4cm}
\caption{\footnotesize
  BRDF reconstruction results on real data. We tried different mobile devices to capture raw images using Adobe LightRoom. The input images in were captured using a Huawei P9 (first three rows), Google Tango (fourth row) and iPhone 6s (fifth row), all with a handheld mobile phone where the z-axis of camera was only approximately perpendicular to the sample surface.}
\label{qualitativeReal}
\vspace{-0.4cm}
\end{figure}

\vspace{-0.3cm}
\paragraph{Qualitative results with different mobile phones}
Figure \ref{qualitativeReal} presents SVBRDF and normal estimation results for real images captured with three different mobile devices: Huawei P9, Google Tango and iPhone 6s. We observe that even with a single image, our network successfully predicts the SVBRDF and normals, with images rendered using the predicted parameters appear very similar to the input. Also, the exact same network generalizes well to different mobile devices, which shows that our data augmentation successfully helps the network factor out variations across devices. For some materials with specular highlights, the network can hallucinate information lost due to saturation. The network can also reconstruct reasonable normals even for complex instances. 

%Please refer to supplementary materials for more results on real data. \KS{Add and reference results with different phone from rebuttal? The arXiv paper can be longer than 8 pages...}\ZL{This experiment is included in appendix, do you think I should move it to the main paper?}

\begin{figure}[!!t]
\centering
\begin{minipage}[c]{0.47\textwidth}
\centering
\setlength{\tabcolsep}{1pt}
\raisebox{-1.8cm}{
\begin{tabular}{ccccc}
\includegraphics[width=0.40in]{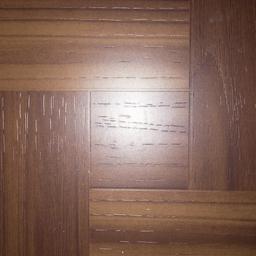} &
\includegraphics[width=0.40in]{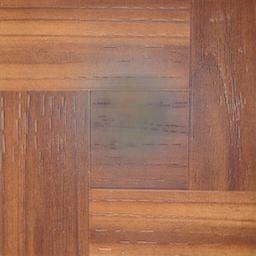} &
\includegraphics[width=0.40in]{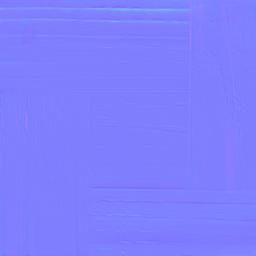} &
\includegraphics[width=0.40in]{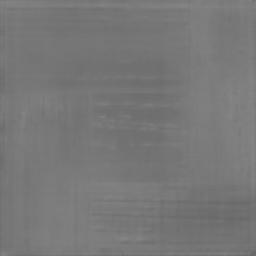} &
\includegraphics[width=0.40in]{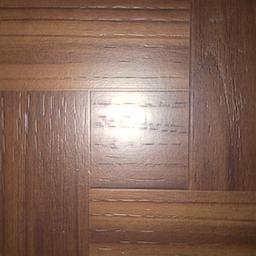} \\
\scriptsize{Input} & \scriptsize{Albedo} & \scriptsize{Normal} & \scriptsize{Roughness} & \scriptsize{Relighting}
\end{tabular}
}
\end{minipage}\hfill
\hspace{-10cm}
\begin{minipage}[r]{0.50\textwidth}
\caption{A failure case, due to incorrect material classification into \emph{metal}, which causes the specularity to be over-smoothed.}
\vspace{-0.7cm}
\label{failurecase}
\end{minipage}
\end{figure}

\vspace{-0.3cm}
\paragraph{A failure case}
In Figure \ref{failurecase}, we show a failure case. Here, the material is misclassified as metal which causes the specular highlight in the center of image to be over-suppressed. 
%A more robust material classification network may solve this problem. 
In future work, we may address this with more robust material classification, potentially exploiting datasets like \cite{minc}.
%material classification datasets \cite{minc}.

\begin{figure}[!!t]
\vspace{-0.4cm}
\centering
\begin{minipage}[c]{0.50\textwidth}
\includegraphics[width=2.3in,height=1.2in]{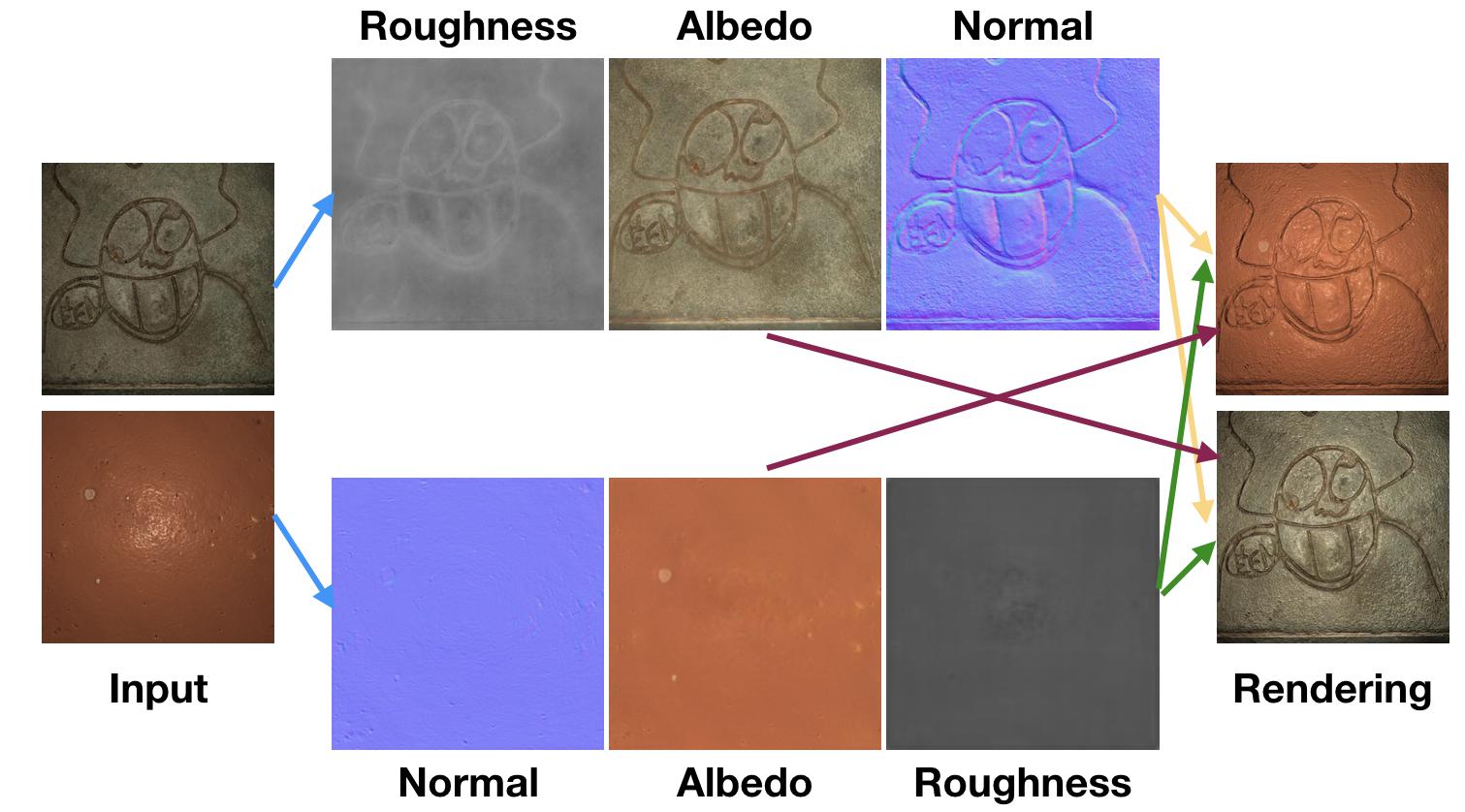}
\end{minipage}\hfill
\begin{minipage}[c]{0.47\textwidth}
\caption{A material editing example. Having reconstructed the SVBRDF and normals of the two samples, we swap the original geometry and material properties, then relight under novel illumination.}
\label{materialEditing}
\end{minipage}
\vspace{-0.2cm}
\end{figure}

\vspace{-0.3cm}
\paragraph{Material editing}
We can edit the reconstructed SVBRDFs by transferring material properties. Figure \ref{materialEditing} shows an example where we transfer BRDF properties across different material types and render in a novel lighting condition.

\vspace{-0.3cm}
\subsection{Further Comparisons with Prior Works}
\begin{figure}[!!t]
\vspace{-0.1cm}
\centering
\begin{minipage}[c]{0.57\textwidth}
\begin{tabular}{cccc|c}
\scriptsize{Our} & \scriptsize{Our} & \scriptsize{\cite{CNN-BRDF} input} & \scriptsize{\cite{CNN-BRDF} normals} & \scriptsize{PS normals} \\[-0.2cm]
\scriptsize{input} & \scriptsize{normals} &  &  &  \\ 
\includegraphics[width=0.45in]{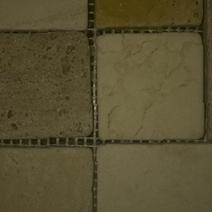} & 
\includegraphics[width=0.45in]{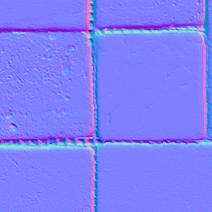} & 
\includegraphics[width=0.45in]{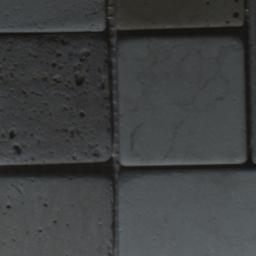} & 
\includegraphics[width=0.45in]{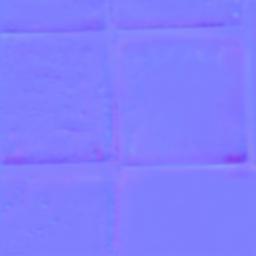} &
\includegraphics[width=0.45in]{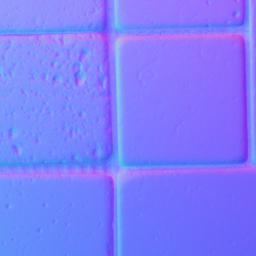} \\
\includegraphics[width=0.45in]{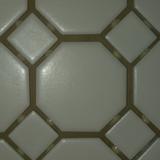} & 
\includegraphics[width=0.45in]{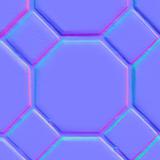} & 
\includegraphics[width=0.45in]{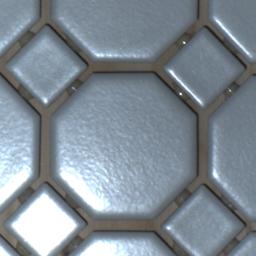} & 
\includegraphics[width=0.45in]{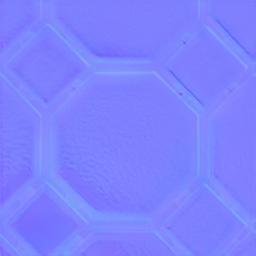} & 
\includegraphics[width=0.45in]{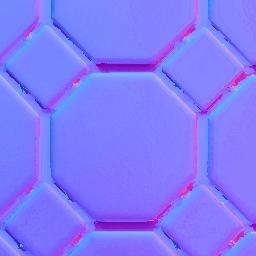}
\end{tabular}
\caption{Comparison of normal maps using our method and \cite{CNN-BRDF}, with photometric stereo as reference. Even with a lightweight acquisition system, our network predicts high quality normal maps.}
\label{normalMap}
\vspace{-0.1cm}
\end{minipage}\hfill
\begin{minipage}[c]{0.40\textwidth}
\begin{tabular}{ccccc}
\includegraphics[width=0.45in]{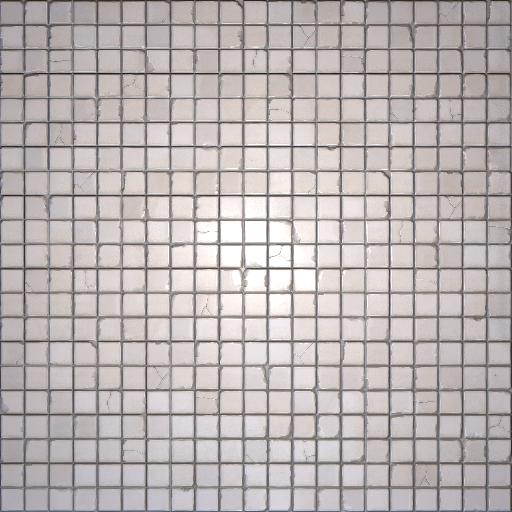} &
\rotatebox[origin=c]{90}{\scriptsize{\qquad\qquad Normal}} &
\includegraphics[width=0.45in]{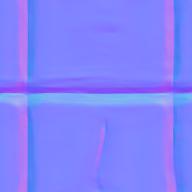} &
\includegraphics[width=0.45in]{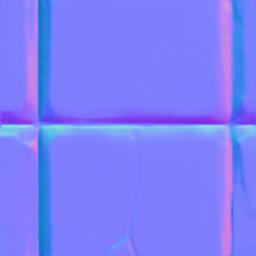} & 
\includegraphics[width=0.45in]{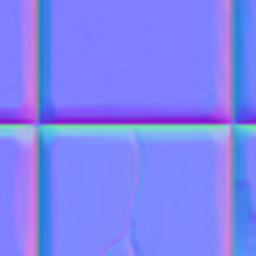} 
\vspace{-0.9cm} \\
\scriptsize{Flash} & & \scriptsize{\cite{twoshotBRDF}} & \scriptsize{Ours} & \scriptsize{GT} \\
\includegraphics[width=0.45in]{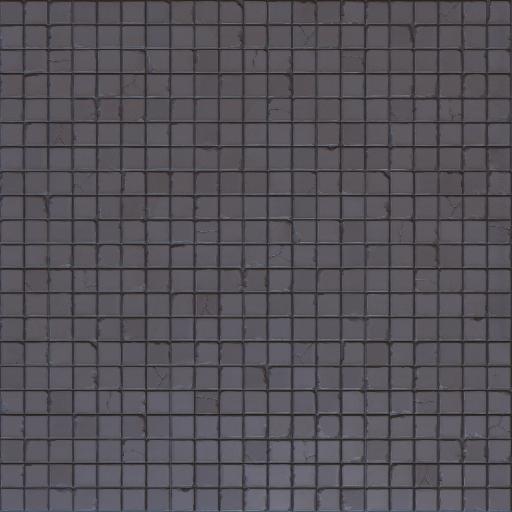} &
\rotatebox[origin=c]{90}{\scriptsize{\qquad\qquad Diffuse}} &
\includegraphics[width=0.45in]{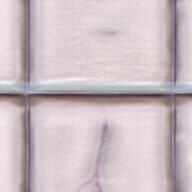} &
\includegraphics[width=0.45in]{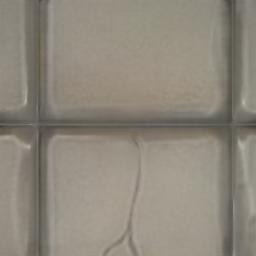} & 
\includegraphics[width=0.45in]{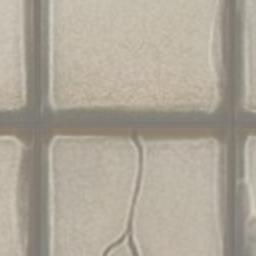} 
\vspace{-0.9cm}\\
\scriptsize{Guide} & & \scriptsize{\cite{twoshotBRDF}} & \scriptsize{Ours} & \scriptsize{GT}
\end{tabular}
\caption{Comparison with \cite{twoshotBRDF}, which requires two images, assumes stationary textures and takes over 6 hours (with GPU acceleration), yet our result is more accurate.}
\label{otherMethodsApp}
\end{minipage}
\vspace{-0.5cm}
\end{figure}

\vspace{-0.3cm}
\paragraph{Comparison with two-shot BRDF method \cite{twoshotBRDF}}
%\KS{is this synthetic data? how was it rendered for the two-shot method?}\ZL{This is synthetic data rendered with our BRDF model. The first show is rendered with point light source + environment map while the second is rendered with environment map only.}
The two-shot method of \cite{twoshotBRDF} can only handle images with stationary texture while our method can reconstruct arbitrarily varying SVBRDFs. For a meaningful comparison, in Figure \ref{otherMethodsApp}, we compare our method with \cite{twoshotBRDF} on a rendered stationary texture. We can see that even for this restrictive material type, the normal maps reconstructed by the two methods are quite similar, but the diffuse map reconstructed by our method is closer to ground truth. While \cite{twoshotBRDF} takes about $6$ hours to reconstruct a patch of size $192\times 192$, our method requires $2.4$ seconds. The aligned flash and no-flash pair for \cite{twoshotBRDF} is not trivial to acquire (especially on mobile cameras with effects like rolling shutter), making our single image BRDF estimation more practical.

\vspace{-0.3cm}
\paragraph{Comparison normals with environment light and photometric stereo}
In Figure \ref{normalMap}, we compare our normal map and the output of a single image SVBRDF reconstruction method under environment lighting~\cite{CNN-BRDF} with photometric stereo \cite{hui2015dic}. We observe that the normals reconstructed by our method are of higher quality than \cite{CNN-BRDF}, with details comparable or sharper than photometric stereo. 
%The supplementary material shows further examples.

\vspace{-0.3cm}
\subsubsection*{Appendix}
The appendix provides further experiments and details, including:
\vspace{-0.2cm}
\begin{tight_itemize}
\item Details of data augmentation, continuous DCRF and visualization of weights
%\item Details of continuous DCRF and visualization of weights
%\item Details of microfacet BRDF model and our data augmentation
\item Spherical renderings of estimated real spatially varying BRDFs
%\item Error distributions of albedo, roughness, normals and relighted images
\item Visualization of SVBRDF estimation with respect to prediction error
\item Further qualitative results on synthetic and real data.
\end{tight_itemize}

\vspace{-0.5cm}
\section{Discussion}
\label{sec:discussion}
\vspace{-0.3cm}

We have proposed a framework for acquiring spatially-varying BRDF using a single mobile phone image. Our solution uses a convolutional neural network whose architecture is specifically designed to capture various physical insights into the problem of BRDF estimation. We also propose a dataset that is larger and better-suited to the problem of material estimation as compared to prior ones, as well as simple acquisition settings that are nevertheless effective for SVBRDF estimation. Our network generalizes very well to real data, obtaining high-quality results in unconstrained test environments. A key goal for our work is to take accurate material estimation from expensive and controlled lab setups, into the hands of non-expert users with consumer devices, thereby opening the doors to new applications. Our future work will take the next step of acquiring SVBRDF with unknown shapes, as well as study the role of other semantic signals such as object categories in material estimation.

%\thispagestyle{empty}
%{\small
\bibliographystyle{splncs}
\bibliography{egbib_supplementary}
%}

\appendix
%\vspace{-0.7cm}
\section{Further Experimental Analysis}
\begin{figure}
\centering
\setlength{\tabcolsep}{2pt}
\begin{tabular}{cccc}
\includegraphics[width=1.1in]{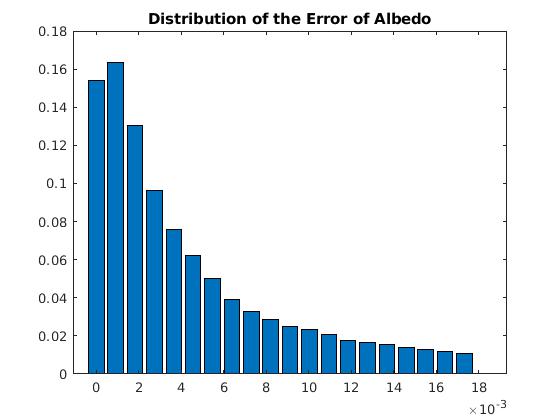} & 
\includegraphics[width=1.1in]{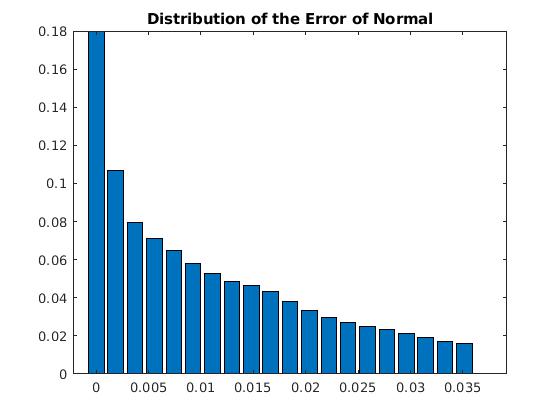} & \includegraphics[width=1.1in]{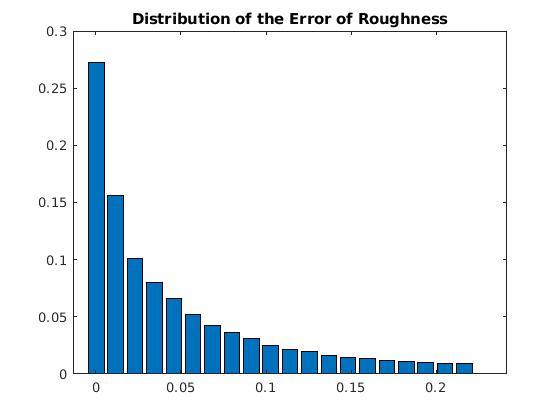} &
\includegraphics[width=1.1in]{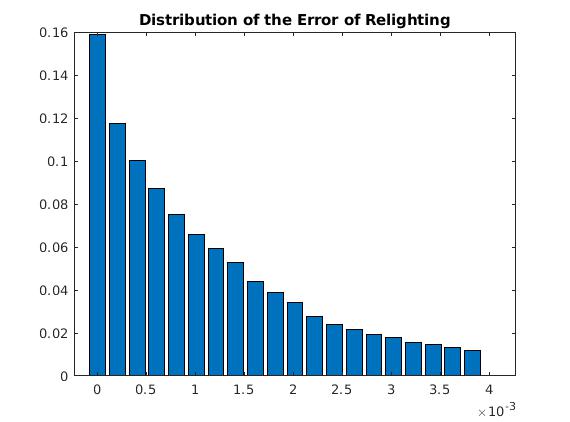}\\
Albedo & Normal & Roughness & Relighting
\end{tabular}
\caption{From the left to the right, error distributions of diffuse albedo, normal, roughness and relighting.}
\label{errorDist}
\end{figure}

\begin{figure*}[!!t]
\centering
\setlength{\tabcolsep}{1pt}
\begin{tabular}{cccccc@{\hspace{0.4cm}}ccccc}
\centering
& \scriptsize{Input} & \scriptsize{Albedo} & \scriptsize{Normals} & \scriptsize{Rough} & \scriptsize{Render}& \scriptsize{Input} & \scriptsize{Albedo} & \scriptsize{Normals} & \scriptsize{Rough} & \scriptsize{Render} \\
\rotatebox[origin=c]{90}{\scriptsize{\qquad\qquad GT}} & 
\includegraphics[width=0.42in]{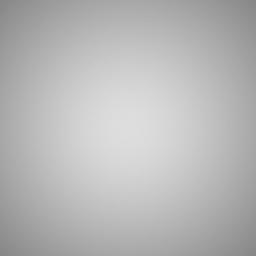} & 
\includegraphics[width=0.42in]{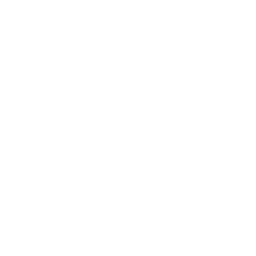} & 
\includegraphics[width=0.42in]{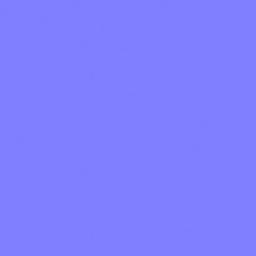} & 
\includegraphics[width=0.42in]{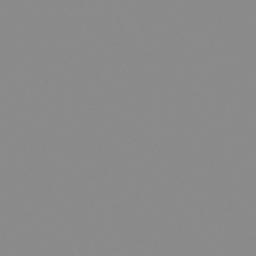} & 
\includegraphics[width=0.42in]{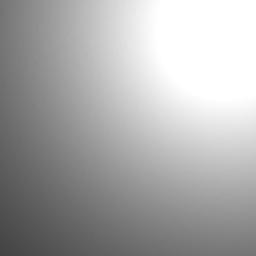} & 
\includegraphics[width=0.42in]{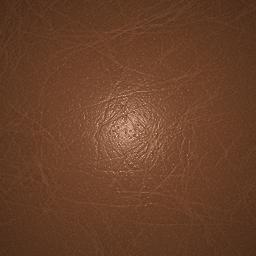} & 
\includegraphics[width=0.42in]{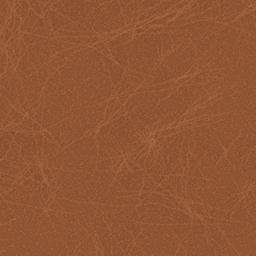} & 
\includegraphics[width=0.42in]{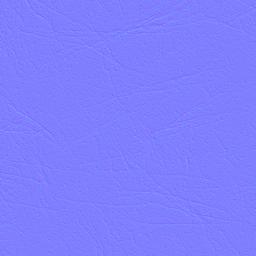} & 
\includegraphics[width=0.42in]{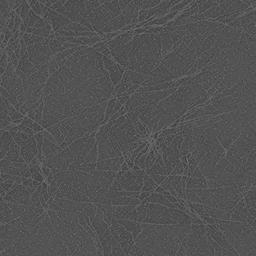} & 
\includegraphics[width=0.42in]{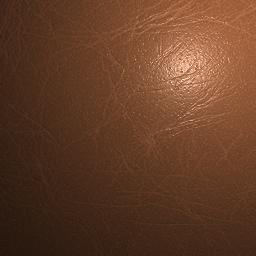} 
\vspace{-0.5cm} \\ 
\rotatebox[origin=c]{90}{\scriptsize{\qquad\qquad Pred}}  & 
\includegraphics[width=0.42in]{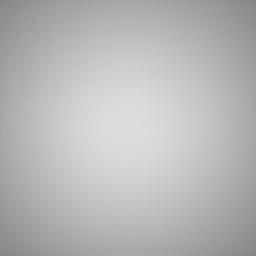} & 
\includegraphics[width=0.42in]{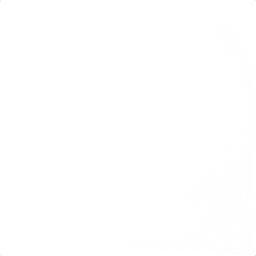} & 
\includegraphics[width=0.42in]{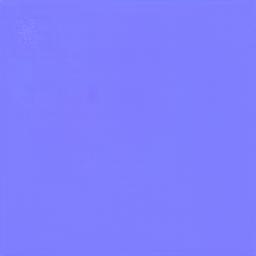} & 
\includegraphics[width=0.42in]{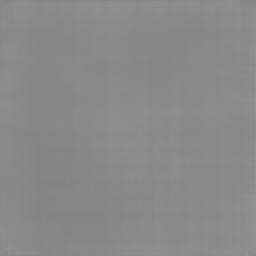} & 
\includegraphics[width=0.42in]{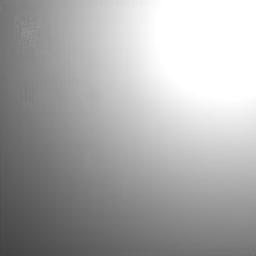} &
\includegraphics[width=0.42in]{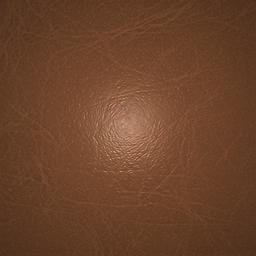} & 
\includegraphics[width=0.42in]{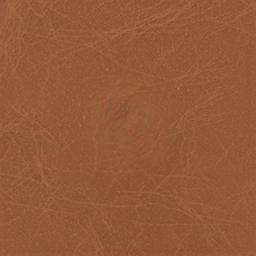} & 
\includegraphics[width=0.42in]{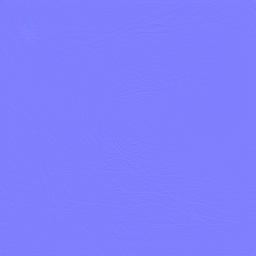} & 
\includegraphics[width=0.42in]{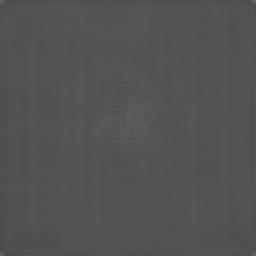} & 
\includegraphics[width=0.42in]{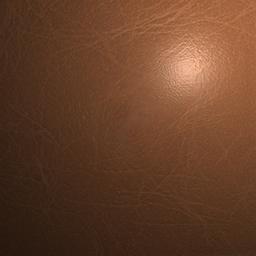} 
\vspace{-0.5cm} \\ 
& \multicolumn{5}{c}{$\mathcal{L}_{d} + \mathcal{L}_{n} + \mathcal{L}_{rec} = 1.7\times 10^{-4}$, $P = 0.0\%$} &  \multicolumn{5}{c}{$\mathcal{L}_{d} + \mathcal{L}_{n} + \mathcal{L}_{rec} = 3.2\times 10^{-3}$, $P = 12.5\%$} \\
\rotatebox[origin=c]{90}{\scriptsize{\qquad\qquad GT}} & 
\includegraphics[width=0.42in]{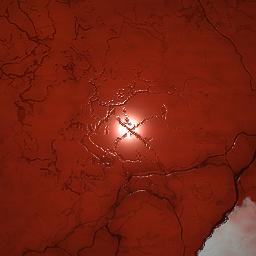} & 
\includegraphics[width=0.42in]{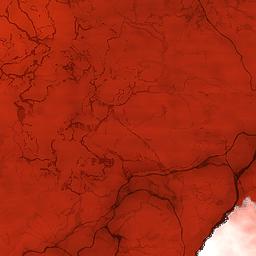} & 
\includegraphics[width=0.42in]{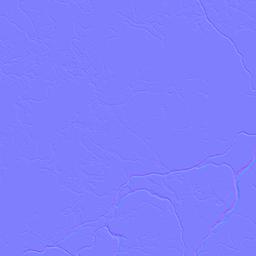} & 
\includegraphics[width=0.42in]{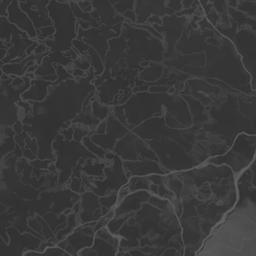} & 
\includegraphics[width=0.42in]{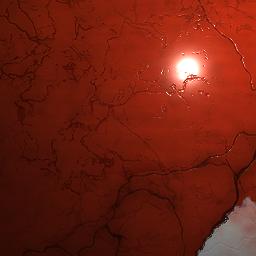} & 
\includegraphics[width=0.42in]{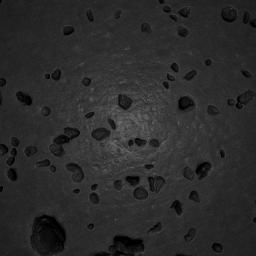} & 
\includegraphics[width=0.42in]{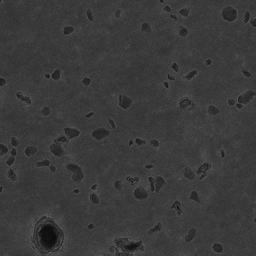} & 
\includegraphics[width=0.42in]{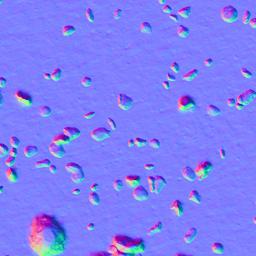} & 
\includegraphics[width=0.42in]{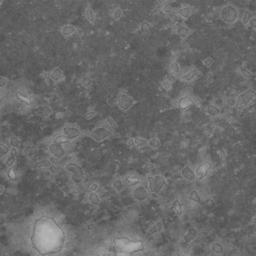} & 
\includegraphics[width=0.42in]{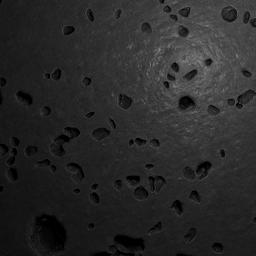} 
\vspace{-0.5cm} \\ 
\rotatebox[origin=c]{90}{\scriptsize{\qquad\qquad Pred}}  & 
\includegraphics[width=0.42in]{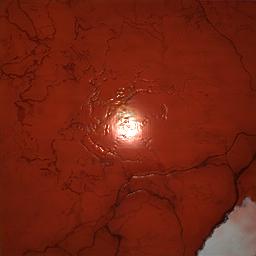} & 
\includegraphics[width=0.42in]{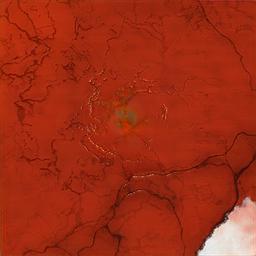} & 
\includegraphics[width=0.42in]{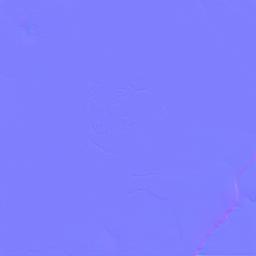} & 
\includegraphics[width=0.42in]{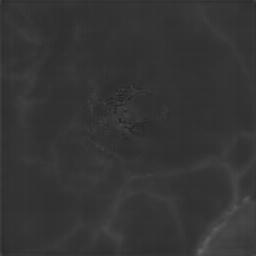} & 
\includegraphics[width=0.42in]{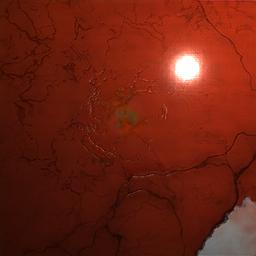} &
\includegraphics[width=0.42in]{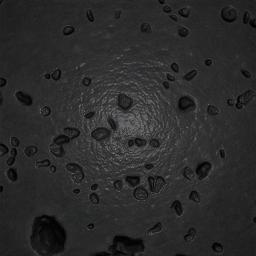} & 
\includegraphics[width=0.42in]{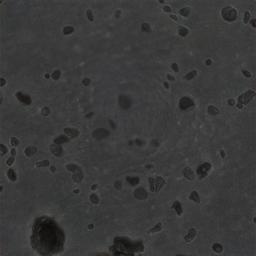} & 
\includegraphics[width=0.42in]{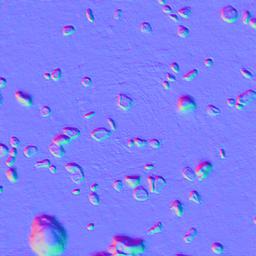} & 
\includegraphics[width=0.42in]{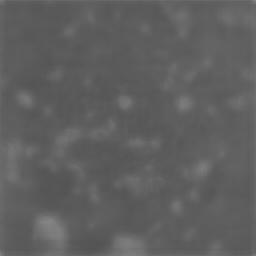} & 
\includegraphics[width=0.42in]{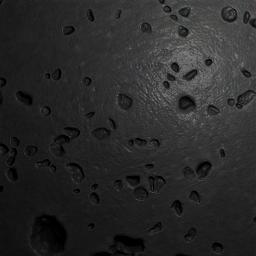} 
\vspace{-0.5cm} \\ 
& \multicolumn{5}{c}{$\mathcal{L}_{d} + \mathcal{L}_{n} + \mathcal{L}_{rec} = 6.4\times 10^{-3}$, $P = 25.0\%$} &  \multicolumn{5}{c}{$\mathcal{L}_{d} + \mathcal{L}_{n} + \mathcal{L}_{rec} = 1.1\times 10^{-2}$, $P = 37.5\%$} \\
\rotatebox[origin=c]{90}{\scriptsize{\qquad\qquad GT}} & 
\includegraphics[width=0.42in]{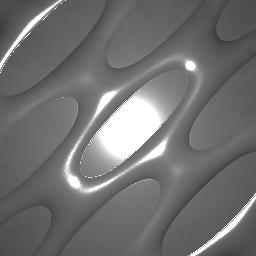} & 
\includegraphics[width=0.42in]{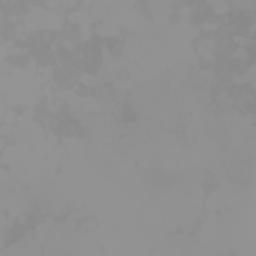} & 
\includegraphics[width=0.42in]{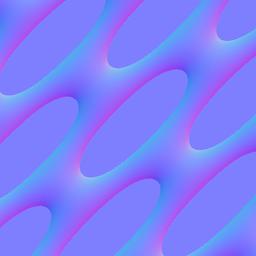} & 
\includegraphics[width=0.42in]{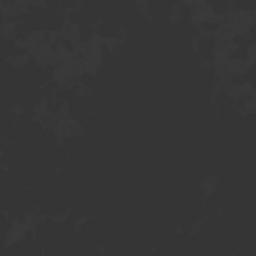} & 
\includegraphics[width=0.42in]{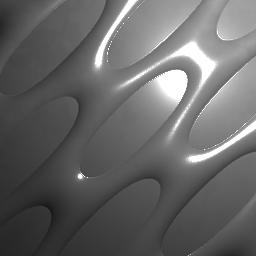} & 
\includegraphics[width=0.42in]{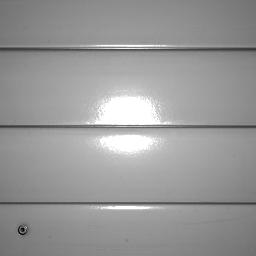} & 
\includegraphics[width=0.42in]{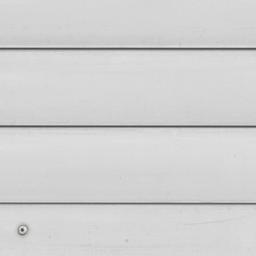} & 
\includegraphics[width=0.42in]{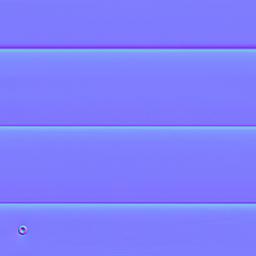} & 
\includegraphics[width=0.42in]{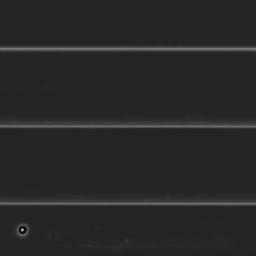} & 
\includegraphics[width=0.42in]{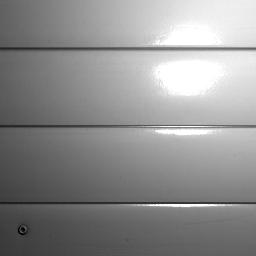} 
\vspace{-0.5cm} \\ 
\rotatebox[origin=c]{90}{\scriptsize{\qquad\qquad Pred}}  & 
\includegraphics[width=0.42in]{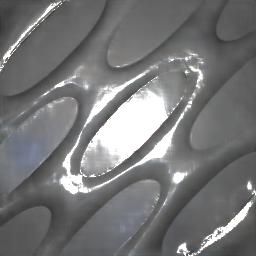} & 
\includegraphics[width=0.42in]{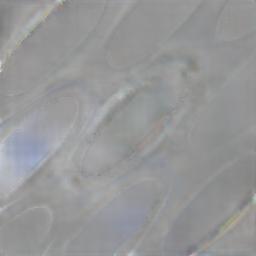} & 
\includegraphics[width=0.42in]{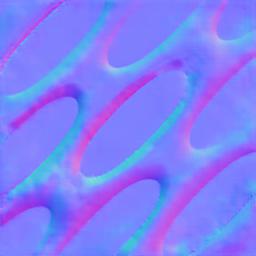} & 
\includegraphics[width=0.42in]{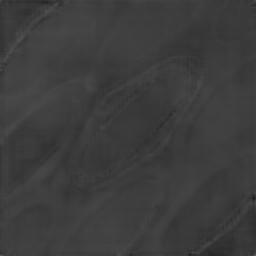} & 
\includegraphics[width=0.42in]{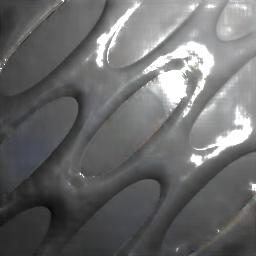} &
\includegraphics[width=0.42in]{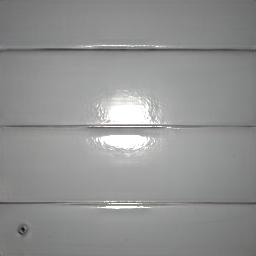} & 
\includegraphics[width=0.42in]{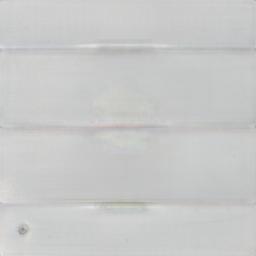} & 
\includegraphics[width=0.42in]{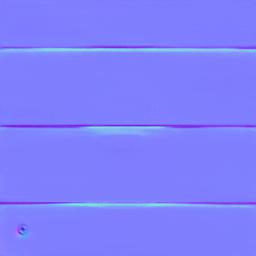} & 
\includegraphics[width=0.42in]{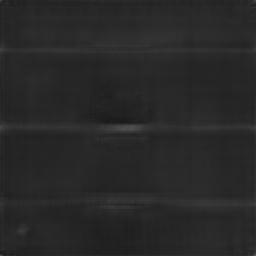} & 
\includegraphics[width=0.42in]{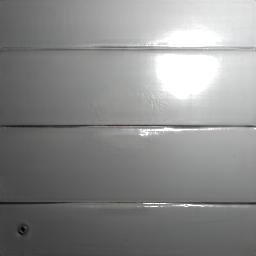} 
\vspace{-0.5cm}\\ 
& \multicolumn{5}{c}{$\mathcal{L}_{d} + \mathcal{L}_{n} + \mathcal{L}_{rec} = 1.7\times 10^{-2}$, $P = 50.0\%$} &  \multicolumn{5}{c}{$\mathcal{L}_{d} + \mathcal{L}_{n} + \mathcal{L}_{rec} = 2.5\times 10^{-2}$, $P = 62.5\%$} \\
\rotatebox[origin=c]{90}{\scriptsize{\qquad\qquad GT}} & 
\includegraphics[width=0.42in]{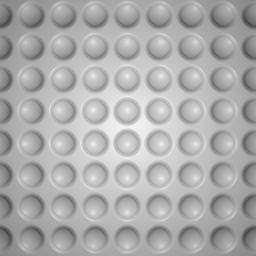} & 
\includegraphics[width=0.42in]{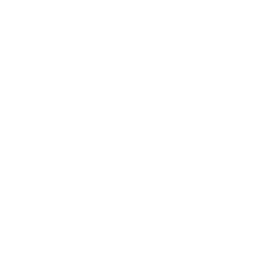} & 
\includegraphics[width=0.42in]{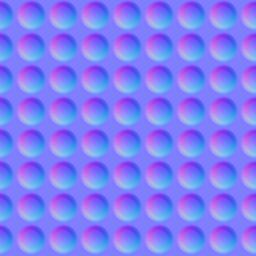} & 
\includegraphics[width=0.42in]{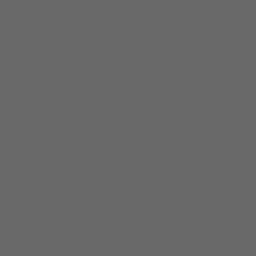} & 
\includegraphics[width=0.42in]{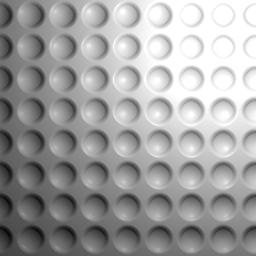} & 
\includegraphics[width=0.42in]{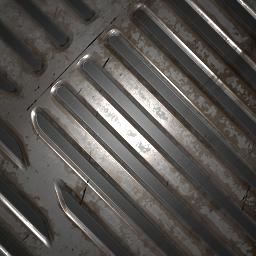} & 
\includegraphics[width=0.42in]{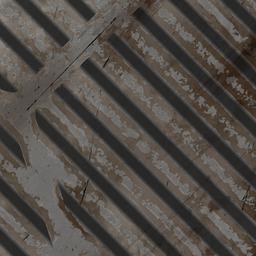} & 
\includegraphics[width=0.42in]{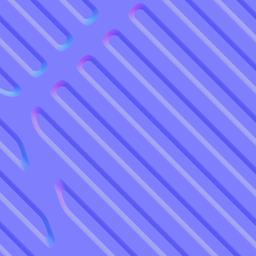} & 
\includegraphics[width=0.42in]{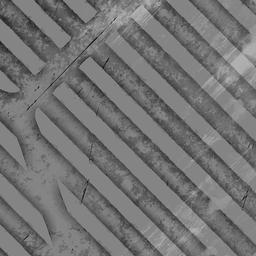} & 
\includegraphics[width=0.42in]{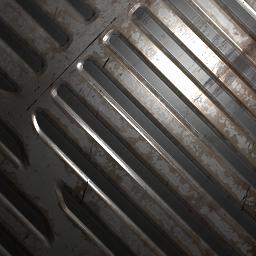} 
\vspace{-0.5cm} \\ 
\rotatebox[origin=c]{90}{\scriptsize{\qquad\qquad Pred}}  & 
\includegraphics[width=0.42in]{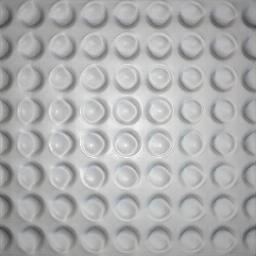} & 
\includegraphics[width=0.42in]{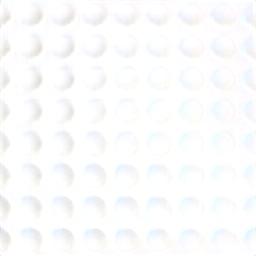} & 
\includegraphics[width=0.42in]{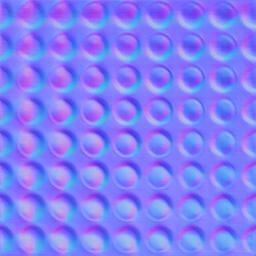} & 
\includegraphics[width=0.42in]{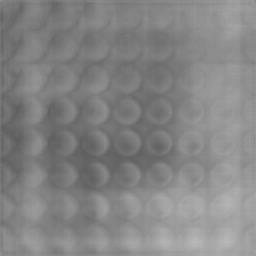} & 
\includegraphics[width=0.42in]{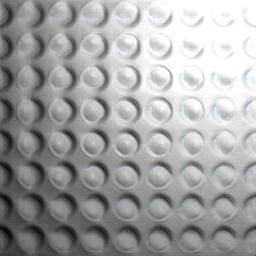} &
\includegraphics[width=0.42in]{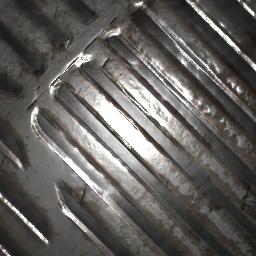} & 
\includegraphics[width=0.42in]{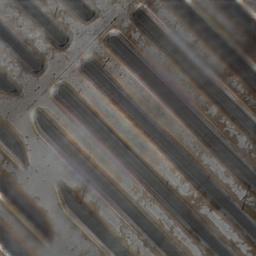} & 
\includegraphics[width=0.42in]{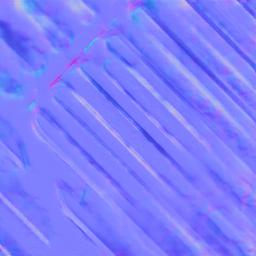} & 
\includegraphics[width=0.42in]{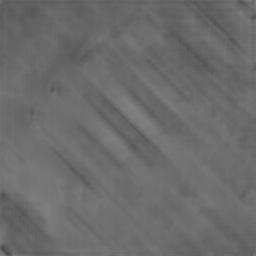} & 
\includegraphics[width=0.42in]{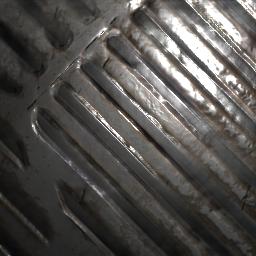} 
\vspace{-0.5cm}\\ 
& \multicolumn{5}{c}{$\mathcal{L}_{d} + \mathcal{L}_{n} + \mathcal{L}_{rec} = 3.7\times 10^{-2}$, $P = 75.0\%$} & \multicolumn{5}{c}{$\mathcal{L}_{d} + \mathcal{L}_{n} + \mathcal{L}_{rec} = 5.2\times 10^{-2}$, $P = 87.5\%$} \\
\end{tabular}
\caption{SVBRDF estimation results sorted according to the prediction error. The error here is defined as $\mathcal{L}_{d} + \mathcal{L}_{n} + \mathcal{L}_{rec}$. We do not consider $\mathcal{L}_{r}$ here roughness has relatively smaller influence towards the final appearance of the surface. Here, $P$ denotes the percentage of samples in the test set with error higher than the considered sample.}
\label{errorExample}
\end{figure*}

\vspace{-0.1cm}
\paragraph{Error distribution on test set}
To provide better intuition into our quantitative results, we plot the distributions of prediction errors for diffuse albedo ($\mathcal{L}_{d}$), normals ($\mathcal{L}_{n}$), roughness ($\mathcal{L}_{r}$) and relighting ($\mathcal{L}_{rec}$) in Figure \ref{errorDist}. Then, we sort the BRDF reconstruction results in the test set according to $\mathcal{L}_{d} + \mathcal{L}_{n} + \mathcal{L}_{rec}$ and illustrate the estimation and relighting quality for a random material picked from various percentiles of the above error distribution. The qualitative comparison is shown in Figure \ref{errorExample}. 

Even though our network is trained end-to-end, we observe physically meaningful trends in Figure \ref{errorDist}. For instance, the materials that correspond to lower error percentiles tend to have flat normals, uniform diffuse color and wide specular lobes. On the other hand, materials with higher errors tend to have more complex normals, stronger local variations in diffuse color and roughness, or more prominent highlights. This demonstrates the benefits of our network design which considers the underlying problem structure. We also observe that normal and diffuse color estimates are quite accurate even at error percentiles higher than $50$, which contributes to reasonable relighting results under \emph{novel} lighting even at high error percentiles.

\begin{figure}
\centering
\setlength{\tabcolsep}{2pt}
\begin{tabular}{cccc|c}
\small{Our Input} & \small{Our Normals} & \small{\cite{CNN-BRDF} Input} & \small{\cite{CNN-BRDF} Normals} & \small{PS Normals} \\ 
\includegraphics[width=0.6in]{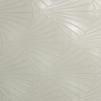} & 
\includegraphics[width=0.6in]{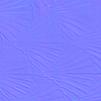} & 
\includegraphics[width=0.6in]{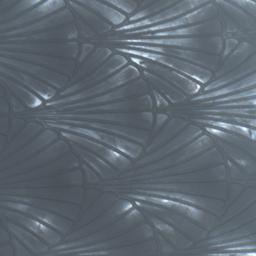} & 
\includegraphics[width=0.6in]{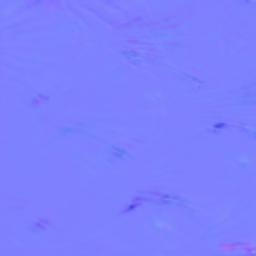} &
\includegraphics[width=0.6in]{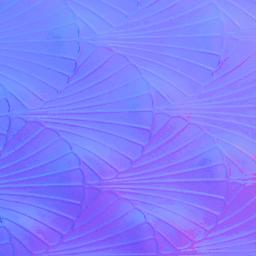} \\
\includegraphics[width=0.6in]{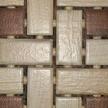} & 
\includegraphics[width=0.6in]{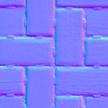} & 
\includegraphics[width=0.6in]{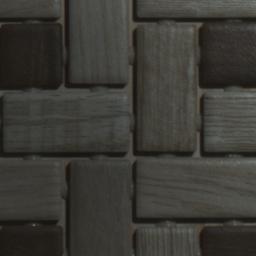} & 
\includegraphics[width=0.6in]{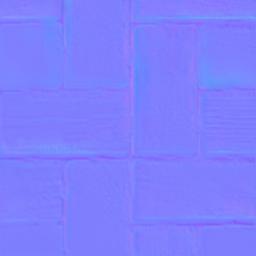} &
\includegraphics[width=0.6in]{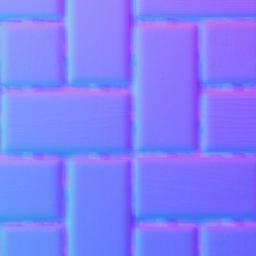} \\
\end{tabular}
\caption{Comparison of normal maps using our method and \cite{CNN-BRDF}, with photometric stereo as reference. Even with a lightweight acquisition system, our network predicts high quality normal maps.}
\begin{comment}
\KS{The GT normals in the fourth row look weird, why is there a curve towards the top right?}\ZL{That's because the image is actually under point light source with projective camera. So the plane will actually be captured as a curve surface. That's the limitation of the PS method. Previously we handled the problem by cropping a small patch from the reconstructed surface. But as you have observed. The plane still seems to be curved.} 
\end{comment}
\label{normalMap_sup}
\vspace{-0.1cm}
\end{figure}

\section{Further Results on Real Data}
\begin{figure*}
\centering
\setlength{\tabcolsep}{1pt}
\begin{tabular}{ccccc @{\hspace{0.2cm}} ccccc}
\centering
 \scriptsize{Input} & \scriptsize{Albedo} & \scriptsize{Normals} & \scriptsize{Roughness} & \scriptsize{Rendering}& \scriptsize{Input} & \scriptsize{Albedo} & \scriptsize{Normals} & \scriptsize{Roughness} & \scriptsize{Rendering} \\
 \includegraphics[width=0.44in]{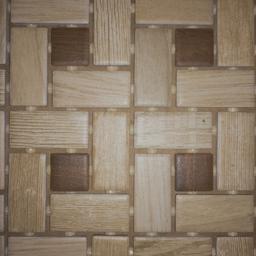} & 
\includegraphics[width=0.44in]{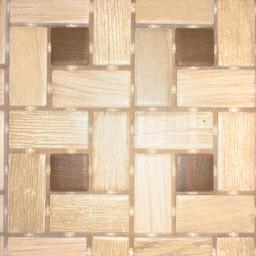} & 
\includegraphics[width=0.44in]{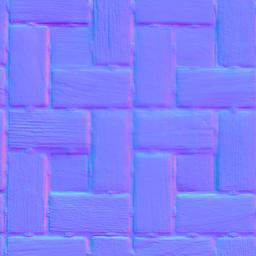} & 
\includegraphics[width=0.44in]{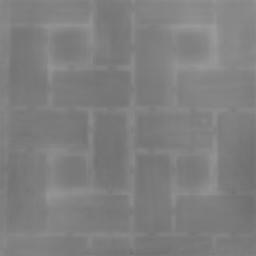} &
\includegraphics[width=0.44in]{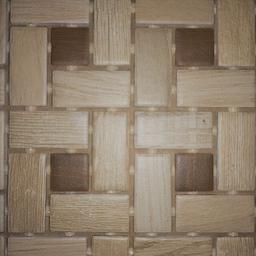} & 
\includegraphics[width=0.44in]{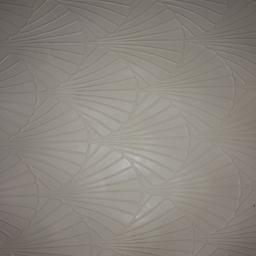} & 
\includegraphics[width=0.44in]{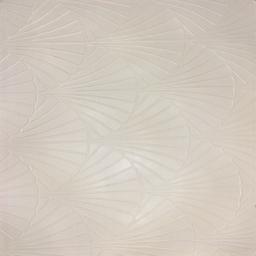} & 
\includegraphics[width=0.44in]{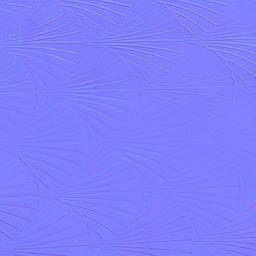} & 
\includegraphics[width=0.44in]{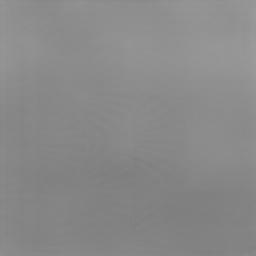} & 
\includegraphics[width=0.44in]{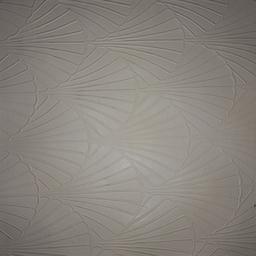} \\
\hline \vspace{-0.2cm}\\
\includegraphics[width=0.44in]{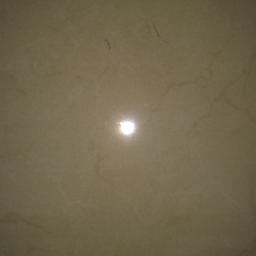} & 
\includegraphics[width=0.44in]{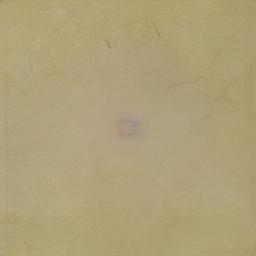} & 
\includegraphics[width=0.44in]{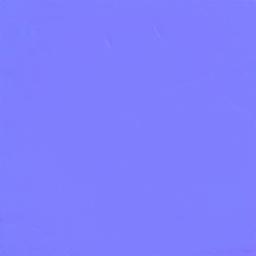} & 
\includegraphics[width=0.44in]{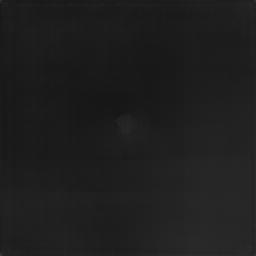} & 
\includegraphics[width=0.44in]{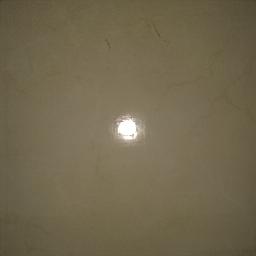} & 
\includegraphics[width=0.44in]{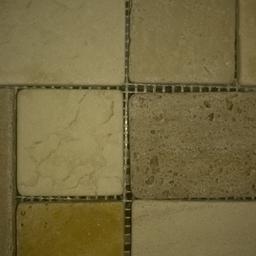} & 
\includegraphics[width=0.44in]{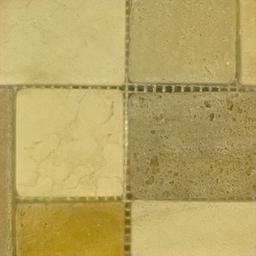} & 
\includegraphics[width=0.44in]{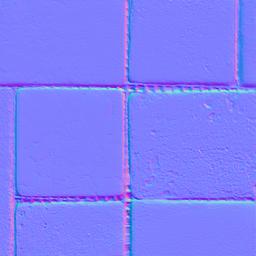} & 
\includegraphics[width=0.44in]{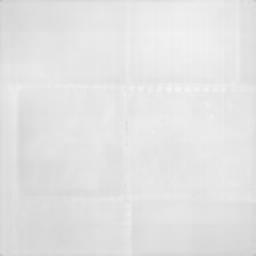} & 
\includegraphics[width=0.44in]{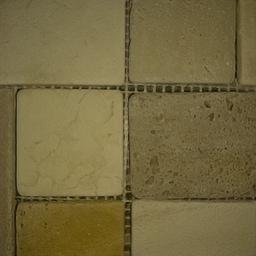} \\
\hline \vspace{-0.2cm} \\
 \includegraphics[width=0.44in]{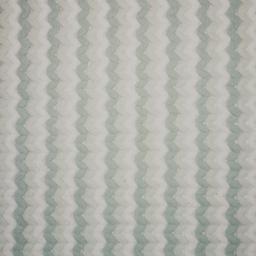} & 
\includegraphics[width=0.44in]{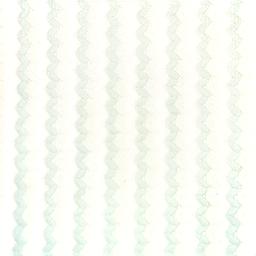} & 
\includegraphics[width=0.44in]{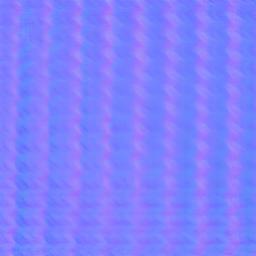} & 
\includegraphics[width=0.44in]{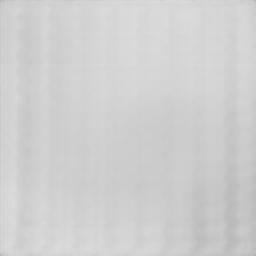} & 
\includegraphics[width=0.44in]{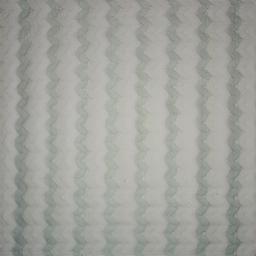} & 
\includegraphics[width=0.44in]{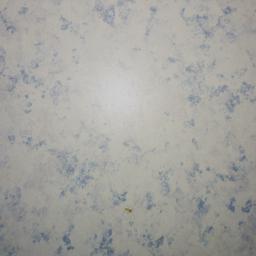} & 
\includegraphics[width=0.44in]{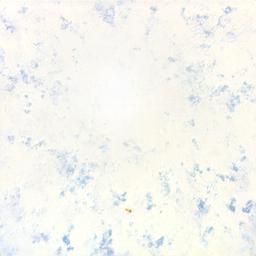} & 
\includegraphics[width=0.44in]{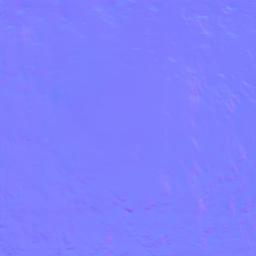} & 
\includegraphics[width=0.44in]{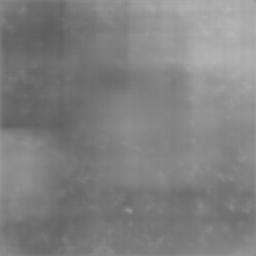} & 
\includegraphics[width=0.44in]{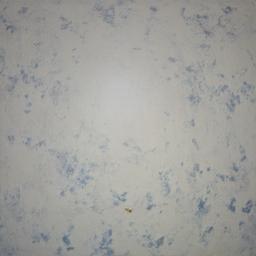} \\
 \includegraphics[width=0.44in]{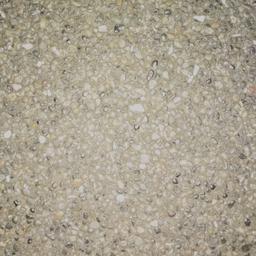} & 
\includegraphics[width=0.44in]{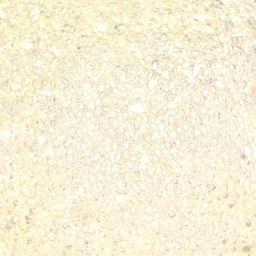} & 
\includegraphics[width=0.44in]{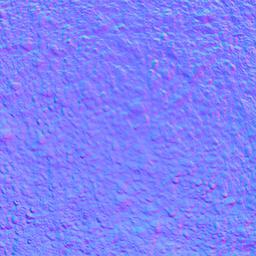} & 
\includegraphics[width=0.44in]{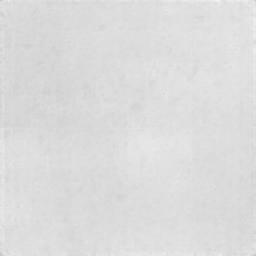} & 
\includegraphics[width=0.44in]{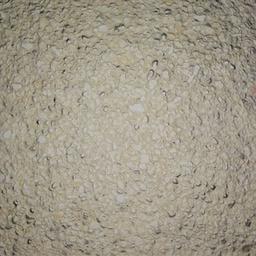} & 
\includegraphics[width=0.44in]{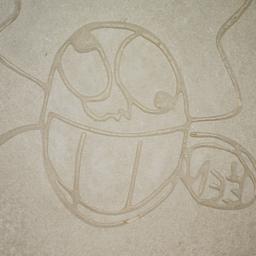} & 
\includegraphics[width=0.44in]{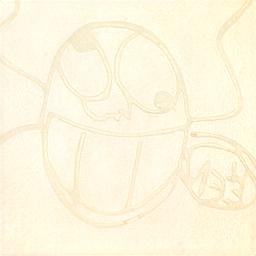} & 
\includegraphics[width=0.44in]{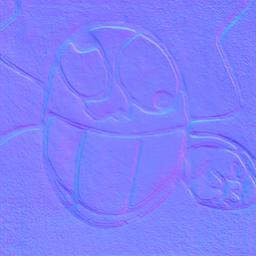} & 
\includegraphics[width=0.44in]{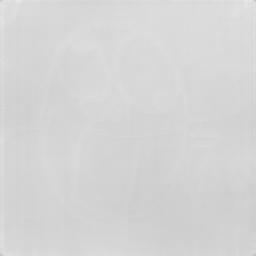} & 
\includegraphics[width=0.44in]{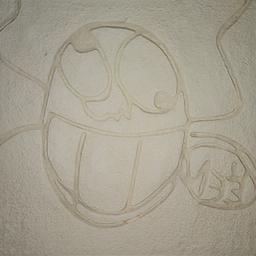} \\
 \includegraphics[width=0.44in]{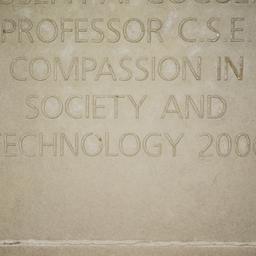} & 
\includegraphics[width=0.44in]{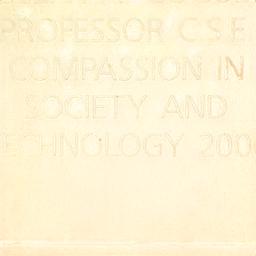} & 
\includegraphics[width=0.44in]{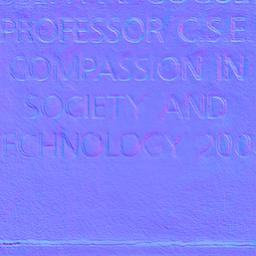} & 
\includegraphics[width=0.44in]{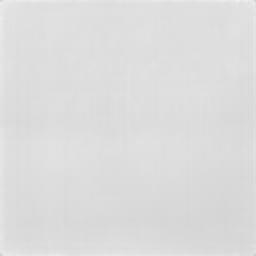} & 
\includegraphics[width=0.44in]{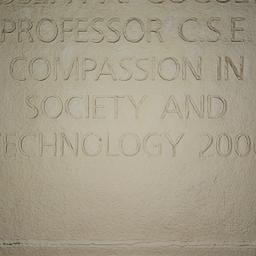} & 
\includegraphics[width=0.44in]{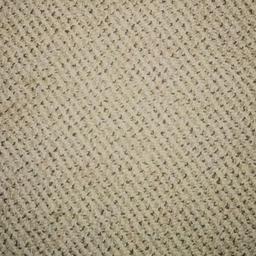} & 
\includegraphics[width=0.44in]{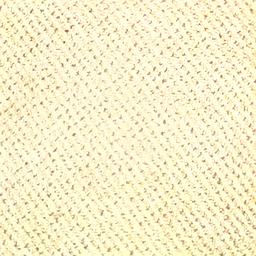} & 
\includegraphics[width=0.44in]{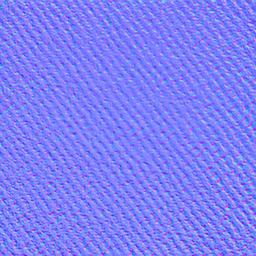} & 
\includegraphics[width=0.44in]{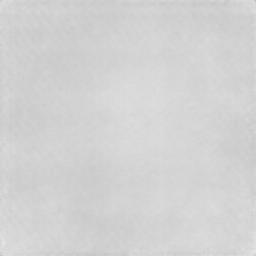} & 
\includegraphics[width=0.44in]{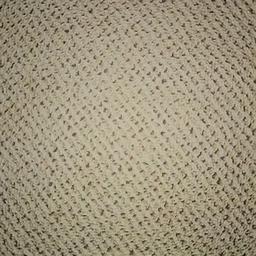} \\
\end{tabular}

\caption{\small
SVBRDF estimation results on real data. All the input images are captured using a mobile phone. All the rows are imaged with a handheld mobile phone, where the z-axis of the camera is only approximately perpendicular to the sample surface. The inaccuracy in positional calibration of the camera is visible in the input image of the example in the second row of the first column, where the highlight is clearly not in the center of the image. However, our method still obtains reasonable normal and SVBRDF estimation results in all cases. The images in the first row are captured by iPhone 6s. The second row are captured by Huawei P9 while the last third rows are captured by Lenovo Phab 2. Our algorithm can handle new unknown devices very well.}
\label{qualitativeReal_sup}
\vspace{-0.1cm}
\end{figure*}

\begin{figure}
\centering
\setlength{\tabcolsep}{1pt}
\begin{tabular}{ccccc}
Input & Albedo & Normal & Roughness & Sphere \\
\includegraphics[width = 0.62in]{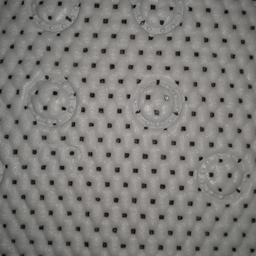} &
\includegraphics[width = 0.62in]{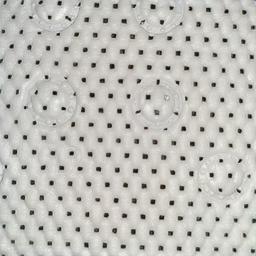} &
\includegraphics[width = 0.62in]{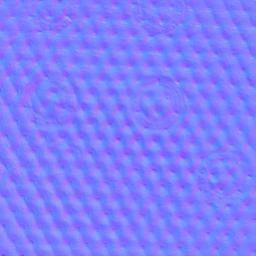} &
\includegraphics[width = 0.62in]{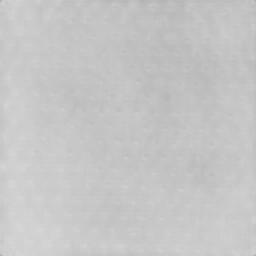} &
\includegraphics[width = 0.62in]{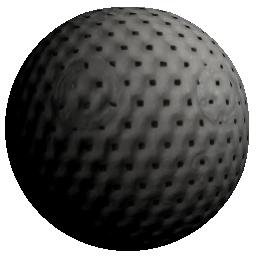} \\
\includegraphics[width = 0.62in]{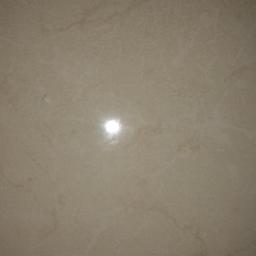} &
\includegraphics[width = 0.62in]{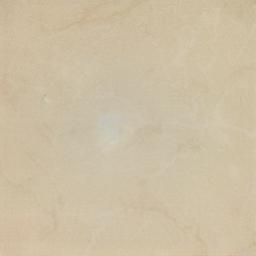} &
\includegraphics[width = 0.62in]{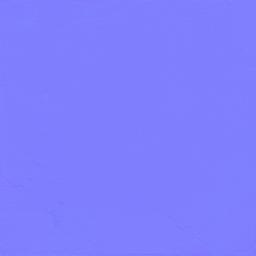} &
\includegraphics[width = 0.62in]{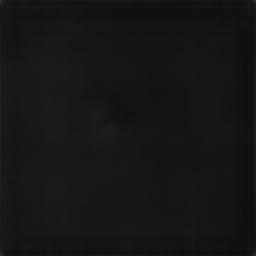} &
\includegraphics[width = 0.62in]{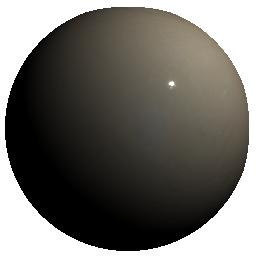} 
\vspace{0.1cm}
\end{tabular}
\caption{Rendering of the estimated real spatially-varying BRDF on a sphere, under a very different oblique lighting direction.}
\label{fig:sphere}
\end{figure}

\paragraph{Comparison with photometric stereo as reference}
In Figure \ref{normalMap_sup}, we compare the normals estimated by our method with that of \cite{CNN-BRDF}, using the normal map from photometric stereo as reference. In the main paper, we use the photometric stereo method of \cite{hui2015dic}. Here, we instead use a simpler but more robust method. We acquire images of a material sample under $52$ different directional point light sources. We abandon the 5 most brightest observations and 5 darkest observations and use the rest for a Lambertian photometric stereo. We find such a method to be quite robust to shadows, as well as the effects of complex BRDF such as glossiness or specularity. We observe that our method is able to capture very fine details in the normal map, in particular, better than the method of \cite{CNN-BRDF}. For instance, note the detail within the grooves of the material in the first and third rows. This demonstrates the efficacy of the proposed method for normal and SVBRDF estimation. 

\paragraph{Real data results in unconstrained environments}
In Figure \ref{qualitativeReal_sup}, we show several more examples of surface normal and BRDF estimation with real data using the proposed method. The images are acquired in unconstrained settings with the camera flash enabled, for several different material types derived from wood flooring, tiles, carpets and so on. In all rows, the mobile phone is hand-held and only approximately parallel to the surface.  In each case, we observe that the recovered normals, as well as the diffuse albedo and specular components of the spatially-varying BRDF appear qualitatively correct. In some cases, such as the second row of the first column, we observe that even very tight specular lobes are well-estimated, as evident from the lobe's compactness in the relighted image. The first row is captured by iPhone 6s, the second row by Huawei P9 and the last three rows by Lenovo Phab 2. Even though we never calibrate the mobile phone, our network generalizes very well to the new device. 

\paragraph{Another visualization for relighting}
For another visualization of the normal and BRDF estimation on real data, we render the estimated material on a sphere illuminated under an oblique lighting direction that is very different from the input lighting. Recall that we only use an approximately planar patch of material as input. The BRDF estimation and relighted sphere are illustrated in Figure \ref{fig:sphere}. We observe that the appearance of the sphere even under a novel lighting direction is quite reasonable.

\section{Microfacet BRDF Model}
\begin{figure}
\centering
\includegraphics[width=3.2in]{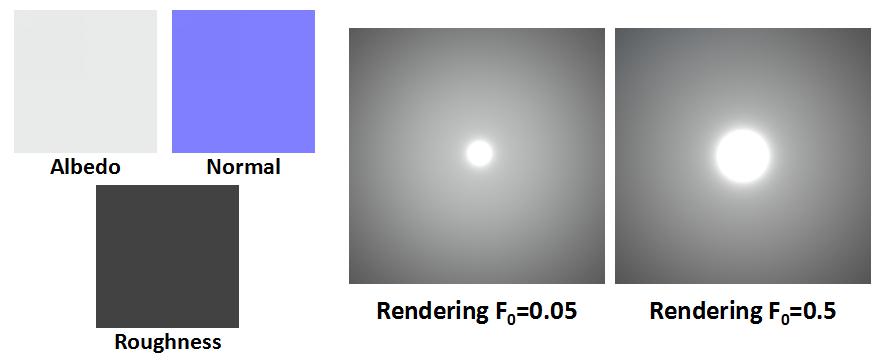}
\caption{An aluminum material rendered with different $F_{0}$. We obersve that when rendering with $F_{0} = 0.5$ the area of specular highlight is much larger and better matches appearnaces of metals in the real world. For all other materials, we use $F_{0} = 0.05$ as the most reasonable value.}
\label{fresnel}
%\vspace{-0.5cm}
\end{figure}

We use the microfacet BRDF model proposed in \cite{brdfModel}. Let $\mathbf{d}_{i}$, $\mathbf{n}_{i}$, $r_{i}$ be the diffuse color, normal and roughness, respectively, at pixel $i$ and $I(\mathbf{d}_{i}, \mathbf{n}_{i}, r_{i})$ be its intensity observed by the camera. Our BRDF model is defined as 
\begin{equation}
\!\!\!\! I(\mathbf{d}_{i}, \mathbf{n}_{i}, r_{i}) = \mathbf{d}_{i} + 
\frac{D(\mathbf{h}_{i}, r_{i})F(\mathbf{v}_{i}, \mathbf{h}_{i})G(\mathbf{l}_{i}, \mathbf{v}_{i}, \mathbf{h}_{i}, r_{i})}{4(\mathbf{n}_{i} \cdot \mathbf{l}_{i})(\mathbf{n}_{i}\cdot \mathbf{v}_{i})} ,
\end{equation}
where $\mathbf{v}_{i}$ and $\mathbf{l}_{i}$ are the view and light directions, while $\mathbf{h}_{i}$ is the half angle vector. Further, $D(\mathbf{h}_{i}, r_{i})$, $F(\mathbf{v}_{i}, \mathbf{h}_{i})$ and $G(\mathbf{l}_{i}, \mathbf{v}_{i}, \mathbf{h}_{i}, r_{i})$ are the distribution, Fresnel and geometric terms, respectively, which are defined as
\begin{align}
\!\!\!\! D(\mathbf{h}_{i}, r_{i}) &=  \frac{\alpha_{i}^{2} }{\pi((\mathbf{n}_{i}\cdot\mathbf{h}_{i})^{2}(\alpha_{i}^{2}-1) + 1)^{2} } \\
\!\!\!\! \alpha_{i} &= r_{i}^{2} \\
\!\!\!\! F(\mathbf{v}, \mathbf{h}) &=  (1 - F_{0})2^{(-5.55473(\mathbf{v}\cdot\mathbf{h}) - 6.98316)\mathbf{v}\cdot\mathbf{h}} + F_0 \\
\!\!\!\! G(\mathbf{l}, \mathbf{v}, \mathbf{n}) &=  G_{1}(\mathbf{v}, \mathbf{n})G_{1}(\mathbf{l}, \mathbf{n}) \\
\!\!\!\! k_{i} &= \frac{(r_{i} + 1)^{2}}{8}\\
\!\!\!\! G_{1}(\mathbf{v}, \mathbf{n})  &=  \frac{\mathbf{n\cdot v} }{(\mathbf{n\cdot v})(1-k) + k}\\
\!\!\!\! G_{1}(\mathbf{l}, \mathbf{n}) &=  \frac{\mathbf{n\cdot l}}{(\mathbf{n\cdot v})(1-k) + k} \: ,
\end{align}
with $F_{0}$ the specular reflectance at normal incidence. For a dielectric material, the value of $F_{0}$ is determined by the index of refraction $\eta$:
\begin{equation}
F_{0} = \frac{(1 - \eta)^{2}}{(1 + \eta)^{2}}.
\end{equation}
For a conductor material, it is determined by the index of refraction $\eta$ and the absorption coefficient $\kappa$:
\begin{equation}
F_{0} = \frac{(1+\eta)^{2} + \kappa^{2}}{(1-\eta)^{2} + \kappa^{2}}.
\end{equation}
When rendering our dataset, we set $F_{0} = 0.5$ for \emph{metal} and $F_{0} = 0.05$ for other kinds of materials. Figure \ref{fresnel} shows an example of smooth aluminum material rendered with $F_{0} = 0.05$ and $F_{0} = 0.5$. We observe that the material rendered with $F_{0} = 0.5$ has a much larger area of specular highlight, which matches appearances of metals in practice. 

\section{Details of Continuous DCRFs}
We use continuous densely connected conditional random fields (DCRFs) for post-processing to remove artifacts caused by saturated highlights and noise in the prediction of the neural network \cite{DCRF,learnCRF}. We customize the DCRFs to better suit our problem of spatially-varying BRDF reconstruction. The distinguishing factor for our DCRF construction is the design of spatially varying weight maps that allow incorporating domain specific knowledge into the CRF inference. In the following, we will discuss the design and the intuition behind the usage of the weight map, as well as the details of training and inference for the DCRF.

\begin{figure}
\centering
\setlength{\tabcolsep}{2pt}
\begin{tabular}{cccccc}
Input & Mask & Input & Mask & Input & Mask \\
\includegraphics[width=0.51in]{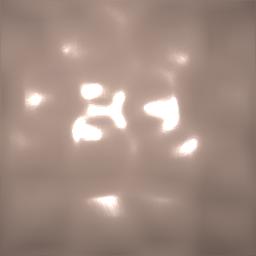} & 
\includegraphics[width=0.51in]{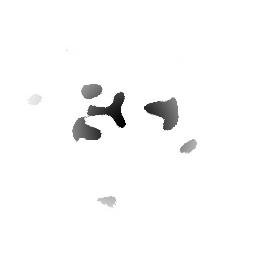} &
\includegraphics[width=0.51in]{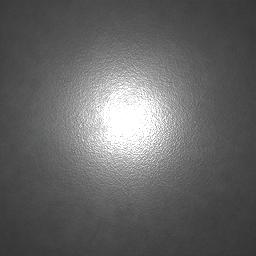} & 
\includegraphics[width=0.51in]{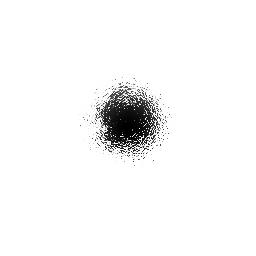} &
\includegraphics[width=0.51in]{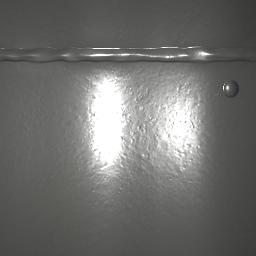} & 
\includegraphics[width=0.51in]{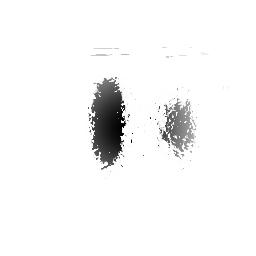} \\
\includegraphics[width=0.51in]{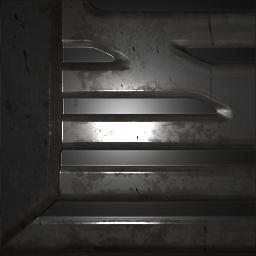} & 
\includegraphics[width=0.51in]{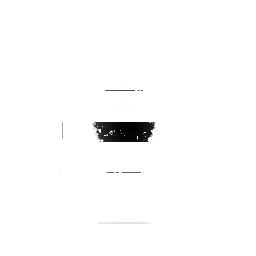} &
\includegraphics[width=0.51in]{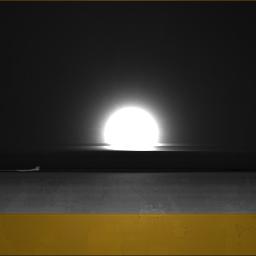} & 
\includegraphics[width=0.51in]{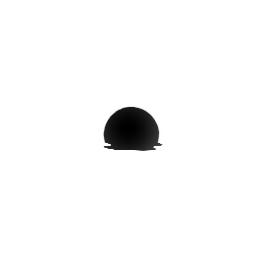} &
\includegraphics[width=0.51in]{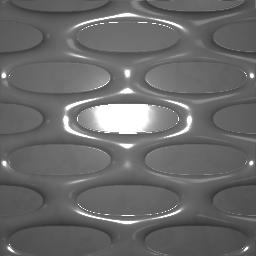} & 
\includegraphics[width=0.51in]{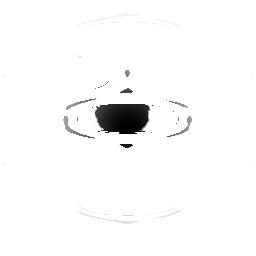} 
\end{tabular}
\caption{The spatially varying weight $\alpha_{i}^{d}$ for the DCRF of diffuse albedo prediction. }
\label{diffuseCRFweightIm}
\end{figure}

\paragraph{Weight Maps of DCRFs}
We first discuss the DCRF for diffuse albedo prediction. Its energy function is defined as 
\begin{eqnarray}
\min_{\{\mathbf{d}_{i}\} }: && \sum_{i=1}^{N} \alpha_{i}^{d}(\mathbf{d}_{i} - \mathbf{\hat{d}}_{i})^{2} + \sum_{i, j}^{N}(\mathbf{d}_{i} - \mathbf{d}_{j})^{2}\Big(\beta_{1}^{d}\kappa_{1}(\mathbf{p}_{i}; \mathbf{p}_{j})  \nonumber\\
&& + \beta_{2}^{d}\kappa_{2}(\mathbf{p}_{i}, \mathbf{\bar{I}}_{i}; \mathbf{p}_{j},\mathbf{\bar{I}}_{j}) + \beta_{3}^{d}\kappa_{3}(\mathbf{p}_{i}, \mathbf{\hat{d}}_{i}; \mathbf{p}_{j},\mathbf{\hat{d}}_{j}) \Big).
\label{diffuseCRFEnergy}
\end{eqnarray}
Here, the coefficient $\alpha_{i}^{d}$ is spatially varying. A larger $\alpha_{i}^{d}$ indicates greater confidence in the prediction from the neural network. Since we use a colocated point light source for illumination, an observation is that saturations caused by the specular highlight are usually in the middle of the image. Another observation is that since the flash illumination is white in color, the saturated pixels are usually white, which means the minimum of their RGB values will be large. Therefore, for regions near the center of the image or regions with specular highlights, we should have a smaller unary weight so that the DCRF may smooth out the artifacts. Based on these two observations, we define the weight map for the unary term $\alpha_{i}^{d}$ as
\begin{eqnarray}
\alpha_{i}^{d} &=& \alpha_{i0}^{d} \max(1 -\exp(-\frac{\mathbf{p}_{i}^{2}}{\sigma_{d0}^{2} }), 1 - \exp(-\frac{(c^{min}_{i} - 1)^{2}}{\sigma_{d1}^{2} } ) ) \nonumber\\
&& + \alpha_{i1}^{d},
\label{diffuseCRFweight}
\end{eqnarray}
where $c^{min}_{i}$ is the minimum of the three color channels at pixel $i$:
\begin{equation}
c^{min}_{i} = \min(R_{i}, G_{i}, B_{i}).
\end{equation}
Here, $\alpha_{i0}^{d}$ and $\alpha_{i1}^{d}$ are two learnable parameters. We set $\alpha_{i1}^{d} = 0$ and $\alpha_{i0}^{d} = 1$ at the beginning of the training process. We set $\sigma_{d0} = 0.5$ and $\sigma_{d1} = 0.08$ through the whole training process. Figure \ref{diffuseCRFweightIm} shows examples of the weight map for diffuse albedo prediction. 

For normal prediction, we do not observe such strong correlation between the prediction error and the position or intensity of the image. Therefore, we just set a uniform weight for every pixel in the image. The energy function is defined as 
\begin{eqnarray}
\min_{\{\mathbf{n}_{i}\}}: &&  \sum_{i=1}^{N}\alpha^{n}(\mathbf{n}_{i} - \mathbf{\hat{n}}_{i})^{2} + \sum_{i, j}^{N}(\mathbf{n}_{i} - \mathbf{n}_{j})^{2}\Big(\beta_{1}^{n}\kappa_{1}(\mathbf{p}_{i};\mathbf{p}_{j}) \nonumber\\
&& + \beta_{2}^{n}\kappa_{2}(\mathbf{p}_{i}, \Delta\mathbf{d}_{i}; \mathbf{p}_{j}, \Delta\mathbf{d}_{j}) \Big) ,
\label{normalCRFEnergy}
\end{eqnarray}
where $\alpha^n$, $\beta_1^n$ and $\beta_2^n$ are learnable parameters that trade-off relative confidences in the unary, a pairwise smoothness prior and a prior on correlation between normals and albedo boundaries.

Finally, for roughness prediction, the energy function is defined as 
\begin{eqnarray}
\min_{ \{r_{i}\} }:  && \sum_{i=1}^{N}\alpha_{i0}^{r}(r_{i} - \hat{r}_{i})^{2} + \alpha_{i1}^{r}(r_{i} - \tilde{r}_{i})^{2} + \sum_{i,j}^{N} (r_{i} - r_{j})^{2}\nonumber\\
&&  \Big(\beta_{1}\kappa_{1}(\mathbf{p}_{i}; \mathbf{p}_{j}) + \beta_{2}\kappa_{2}(\mathbf{p}_{i}, \mathbf{d}_{i}; \mathbf{p}_{j}, \mathbf{d}_{j}) \Big) ,
\label{roughnessCRFEnergy} 
\end{eqnarray}
where $\hat{r}_{i}$ is the prediction from the network and $\tilde{r}_{i}$ is the prediction from a grid search. We find that the prediction from grid search is usually only accurate near the glossy regions, which means these regions should have a larger $\alpha_{i1}^{r}$. Therefore, we define the weight map to be
\begin{eqnarray}
\alpha_{i1}^{r} &=& \max(\exp(-\frac{\mathbf{p}_{i}^{2} }{\sigma_{r0}^{2} } ), \exp(-\frac{c_{m}^{i} - 1 }{\sigma_{r1}^{2} }) ) ,
\end{eqnarray}
where $\alpha_{i0}^{r}$ is constant across the whole image. Both $\alpha_{i0}^{r}$ and $\alpha_{i1}^{r}$ can be learned through back propagating the gradient. We set $\sigma_{r0} = 0.5$ and $\sigma_{r1} = 0.2$. 

\paragraph{Hyperparameters for Training And Inference}
In order to increase the capacity of the DCRF model, we learn different sets of BRDF parameters for each type of material. During both inference and training time, we average the DCRF coefficients according to the output of our material classifier. Let $\{\theta_{i}\} = \{ \{\alpha_{i}\}, \{\beta_{i}\} \}$ be the DCRF coefficients for one material. To enhance the robustness of our method, we re-parameterize the coefficients as
\begin{equation}
\bar{\theta_{i}} = \frac{\theta_{i} }{\sum_{i} \theta_{i} }.
\end{equation}
We clip the DCRF coefficients to always be positive. We use the Adam optimizer to optimize the coefficients. The learning rate is set to $2\times 10^{-4}$ and we reduce it by half after every 2000 iterations. We adopt the method in \cite{learnCRF} to train our DCRF model. The batch size is set to 32. We train the DCRF for diffuse albedo prediction over 4000 iterations and the DCRF for roughness and normal prediction over 3000 iterations. The standard deviations of Gaussian smooth kernels for the three DCRFs are shown in Table \ref{stdGaussian}. 

\begin{table}
\centering
\renewcommand{\arraystretch}{1.3}
\begin{tabular}{cccc}
\hline
\multicolumn{4}{c}{Gaussian Kernels of DCRF for Diffuse Albedo} \\
\hline
 & $\qquad\mathbf{p}_{i}\qquad$ &  $\qquad\mathbf{\bar{I}}_{i}\qquad$    &  $\mathbf{d}_{i}$ \\
$\kappa_{1}$ &0.04 & - & - \\
$\kappa_{2}$ &0.06 & 0.2 & - \\
$\kappa_{3}$ &0.06 & - & 0.1\\
\hline
\end{tabular}
\vspace{0.5cm}

\begin{tabular}{ccc}
\hline
\multicolumn{3}{c}{Gaussian Kernels of DCRF for Normal Map} \\
\hline
 & $\quad\qquad\mathbf{p}_{i}\quad\qquad$ &  $\qquad\Delta\mathbf{d}_{i}\qquad$   \\
$\kappa_{1}$ &0.03 & - \\
$\kappa_{2}$ &0.06 &0.1  \\
\hline
\end{tabular}
\vspace{0.5cm}

\begin{tabular}{ccc}
\hline
\multicolumn{3}{c}{Gaussian Kernels of DCRF for Roughness Map} \\
\hline
 & $\quad\qquad\mathbf{p}_{i}\quad\qquad$ &  $\qquad\mathbf{d}_{i}\qquad$   \\
$\kappa_{1}$ &0.04 & -  \\
$\kappa_{2}$ &0.06 &0.2  \\
\hline
\end{tabular}
\vspace{0.1cm}
\caption{Standard deviations of the Gaussian smoothing kernels of the DCRFs for diffuse albedo, normal and roughness prediction.}
\label{stdGaussian}
\end{table}

\section{Details of Data Augmentation}
In experiments, besides rotating and cropping the original high resolution spatially-varying materials, another important data augmentation is to scale the BRDF parameters for each patch before rendering them into images. For diffuse albedo, we uniformly sample scale coefficients in the range $0.8$ to $1.4$. For normal map, we sample the scale coefficients in the same way, apply the coefficients to the $x$ and $y$ components, then normalize the normal vector to be of unit length. For roughness, we sample the scale coefficients from a Gaussian distribution centered at $1$, with standard deviation equal to $0.2$. Empirically, we observe that such data augmentation can greatly improve the generalization ability of the network. For example, simply scaling the roughness parameter for each patch decreases the validation error for roughness prediction by $15\%$.

 \end{document}